\documentclass{article}

\PassOptionsToPackage{numbers,sort&compress}{natbib}

\usepackage[final]{neurips_2021}

\usepackage[utf8]{inputenc} 
\usepackage[T1]{fontenc}    
\usepackage{hyperref}       
\usepackage{url}            
\usepackage{booktabs}       
\usepackage{amsfonts}       
\usepackage{nicefrac}       
\usepackage{microtype}      
\usepackage{xcolor}         
\usepackage{amsthm}
\usepackage{multirow}
\usepackage{amsmath}
\usepackage{amssymb}
\usepackage{graphicx, subfigure}
\usepackage{caption}

\newcommand{\E}[1]{\mathbb{E}\left[{#1}\right]}
\newcommand{\norm}[1]{\left\vert \left\vert #1 \right\vert \right\vert}
\newcommand{\abs}[1]{\lvert#1\rvert}
\newcommand{\Var}[1]{\mathbf{Var}\left[{#1}\right]}
\DeclareMathOperator{\Tr}{Tr}

\newtheorem{theorem}{Theorem}

\newtheorem{proposition}{Proposition}
\newtheorem{theorem_sup}{Theorem}
\newtheorem*{theorem_sup*}{Theorem}
\newtheorem{lemma_sup}{Lemma}
\newtheorem{proposition_sup}{Proposition}
\newtheorem*{*theorem}{Theorem}
\newtheorem*{remark}{Remark}
\newtheorem*{*corollary}{Corollary}

\title{Faster Neural Network Training  with Approximate Tensor Operations}

\author{%
  Menachem Adelman \\
  Intel \& Technion \\
  \texttt{adelman.menachem@gmail.com} \\
  \And
  Kfir Y. Levy \thanks{A Viterbi fellow}\\
  Technion   \\
  \texttt{kfirylevy@technion.ac.il} \\
  \And
  Ido Hakimi \\
  Technion \\
  \texttt{idohakimi@gmail.com} \\
  \And
  Mark Silberstein \\
  Technion  \\
  \texttt{mark@ee.technion.ac.il} \\
}

\begin{document}

\maketitle

\begin{abstract}
We propose a novel technique for faster deep neural network training which systematically applies sample-based approximation to the constituent tensor operations, i.e., matrix multiplications and convolutions. We introduce new sampling techniques, study their theoretical properties, and prove that they provide the same convergence guarantees when applied to SGD training. 
We apply approximate tensor operations to single and multi-node training of MLP and CNN networks on MNIST, CIFAR-10 and ImageNet datasets. We demonstrate up to 66\% reduction in the amount of computations and communication, and up to 1.37x faster training time while maintaining negligible or no impact on the final test accuracy.
\end{abstract}

\section{Introduction}
Approximation techniques for faster inference and training of deep neural networks (DNNs) have received considerable attention. Examples include  quantization~ \citep{hubara2016quantized,micikevicius2017mixed,seide20141,wen2017terngrad, chakrabarti2019backprop}, low-rank and structured-sparse models \citep{mamalet2012simplifying, kuchaiev2017factorization, mishra2021accelerating, hubara2021accelerated}, 
weight extrapolations \citep{kamarthi1999accelerating}, and partial/asynchronous gradient updates in the context of distributed training~ \citep{recht2011hogwild, strom2015scalable}.
Sampling-based approximations were used to accelerate inference \citep{hua2019channel, gao2018dynamic}, but using them in training \citep{sun2017meprop,chen2019slide,goli2020resprop} has not been systematically studied nor demonstrated end-to-end GPU performance benefits in practice. 

We propose a novel approach to accelerating \emph{DNN training} by systematically approximating tensor operations via sampling. At a high level, the original matrix products and convolutions 
are replaced with their faster approximate versions. 
The approximation is applied separately to each tensor operation, keeping the network architecture and
tensor dimensions intact, thereby facilitating the adoption of this technique in existing DNN 
training frameworks, potentially in combination with other approximation techniques. Furthermore, when combined with distributed training, our technique allows for seamless reduction in the communication bandwidth and increased performance gains.

We begin by reviewing the plethora of existing methods for approximating matrix multiplication. We compare several known algorithms \citep{cohen1999approximating,drineas2001fast,drineas2006fast,magen2011low,sarlos2006improved,clarkson2009numerical,pagh2013compressed,kutzkov2013deterministic},
and find \emph{column-row sampling (CRS)} \citep{drineas2006fast} to be the
most suitable for approximating matrix multiplications in training.
In order to compute the product of two matrices $A^\top B$, the CRS algorithm
samples the columns of $A^\top$  and the corresponding rows of
$B$ thus constructing smaller matrices which are then multiplied as
usual. This method incurs low sampling overheads and lends itself to an efficient implementation using existing dense matrix product routines.  
CRS minimizes the approximation error for the Frobenius norm of the resulting matrix while keeping the approximation unbiased. 

Sampling-based approximations can be interpreted as a form of Dropout~ \citep{srivastava2014dropout}, and we discuss the similarities and differences between the two. While Dropout aims to prevent overfitting, we focus on approximations as means to accelerate training by reducing the amount of computation.

In this work we aim to answer two main questions. First, can neural networks be trained while using approximate tensor operations? Second, what are the relations between using exact or approximate operations during training?

We start by analyzing the simpler case of linear regression,
where we can derive the effects of approximations in closed
form. We define a new loss function that takes the sampling into account, and observe that the resulting gradients differ from the exact training due to the dependency between sampled features. To this end, we propose a new \emph{Bernoulli-CRS} variant which achieves statistical independence of samples, study its properties, and show that in linear regression it is equivalent to dynamic $L_2$ weight regularization of the original, non-approximate loss.

We then turn to the more general case of non-linear deep neural networks. We show that using sampling-based approximations in the backward pass provides the same convergence guarantees as the exact SGD for bounded weights. The convergence result holds for unbounded weights as well if approximation is applied only to the weight gradients and if the activation functions are bounded. 

We also study a new \emph{TopK-CRS} algorithm which deterministically selects the top-$k$ column-row pairs with the highest norms. We show that this algorithm is equivalent to the minimal mean square error estimator (MMSE) in case column-row pairs are pairwise independent with zero mean.

Last, we generalize matrix product approximation to convolutions and analyze the approximation error to derive the optimal sampling policy. This allows us to apply approximations to training of convolutional neural networks (CNNs). 

We implement our techniques in PyTorch \cite{paszke2019pytorch}~\footnote{\url{https://github.com/acsl-technion/approx}} and evaluate them on several DNN topologies, including MLP and CNN networks on MNIST~ \citep{lecun1998gradient}, Wide ResNet 28-10~ \citep{zagoruyko2016wide} on CIFAR-10~ \citep{krizhevsky2009learning}, and ResNet-50 and ResNet-152~ \citep{he2016deep} on ImageNet~ \citep{russakovsky2015imagenet}. We demonstrate up to 66\% reduction in the number of FLOPs and up to 1.33x faster training time with
little or no degradation in model accuracy. 

We develop another flavor of \emph{TopK-CRS} which samples according to the weight norms only. When sampling the same subset of weights for different workers in a data-parallel setting, our sampling technique enables reducing the amount of gradient communication between workers. Notably, our algorithm is compatible with the standard AllReduce approach used in distributed deep learning. We implement an AllReduce scheme that takes advantage of the smaller gradient footprint and demonstrate 1.09x-1.37x speedup in multi-node training.

Our contributions are as follows:
\begin{itemize}
\item We derive general convergence guarantees for training with approximate tensor operations.
\item We develop novel sampling algorithms and analyze their theoretical properties.
\item We extend sampling-based algorithms to fast approximation of multi-channel convolutions.
\item We show that our approach can reduce the computation, communication and total training time on several popular neural network architectures with little or no accuracy degradation.
\end{itemize} 

\section{Related work}

To the best of our knowledge, we are the first to study the application of sample-based approximations of tensor operations to speed up DNN training. 
However, there have been several prior efforts to accelerate DNN computations via approximation which we survey below. 

Several works accelerate inference through model compression \citep{denton2014exploiting,jaderberg2014speeding,lebedev2014speeding,osawa2017accelerating,gong2014compressing,han2015deep,sun2017training}. 
A large body of work is devoted to quantization and
low-precision datatypes (see for example \citep{hubara2016quantized,micikevicius2017mixed, chakrabarti2019backprop}).
Approximation was used to extrapolate weight values \citep{kamarthi1999accelerating}.
Another approach enforces low-rank or structured-sparse structure on the layers, resulting
in lower computational cost both for training and inference \citep{mamalet2012simplifying, kuchaiev2017factorization, mishra2021accelerating,hubara2021accelerated}. Other works accelerate inference by approximating large matrices as products of lower-ranked ones \citep{choromanski2020rethinking, wang2020linformer} or through locality-sensitive hashing \citep{kitaev2020reformer}.

In the context of distributed training, several works targeted communication bottlenecks by gradient quantization \cite{seide20141, wen2017terngrad}, delayed weight updates \citep{recht2011hogwild, strom2015scalable} and low-rank approximation of the gradient matrix \cite{vogels2019powersgd}. These methods are complementary and compatible with ours.

Sampling-based approximations were used to accelerate inference \citep{hua2019channel, gao2018dynamic}, but using them for training\citep{sun2017meprop,chen2019slide,goli2020resprop} has not been systematically studied nor shown to speed up training on GPUs without accuracy degradation. Sub-sampling whole layers was shown to enable training of very deep CNNs \citep{huang2016deep}.

\subsection{Approximate matrix multiplication}
There are several known algorithms for approximating matrix product. However, only those that meet the following requirements will be effective for DNN training. 
First, the algorithm should apply to dense matrices of arbitrary
dimensions. Second, to reduce training time, the overall multiplication including input transformation should be faster than the original product.
Last, the algorithm should be amenable to efficient implementation on commodity hardware. 

Using these criteria, we consider the following algorithms:

\noindent{\bf Random walk~ \citep{cohen1999approximating}} This algorithm performs random walks on a graph representation of the input
matrices, but is applicable to non-negative matrices only.

\noindent{\bf Random projections~ \citep{sarlos2006improved,clarkson2009numerical,magen2011low}} 
The two matrices to be multiplied are first projected into a lower-dimensional
subspace by a scaled random sign matrix. These algorithms require both input
matrices to be roughly square, otherwise the cost of projection will be similar
to the cost of original product. In DNNs, however, it is common for one dimension to be
smaller than the other. 

\noindent{\bf FFT~ \citep{pagh2013compressed, kutzkov2013deterministic}}
These algorithms represent each column-row outer product as a polynomial multiplication 
and then calculate it using Fast Fourier Transform. The
complexity depends on the sparsity of the input matrices, decreasing as the
sparsity increases. Therefore, these algorithms might not be effective for
dense matrices. 

\noindent{\bf SVD~ \citep{drineas2001fast, denton2014exploiting,
osawa2017accelerating}}
Several algorithms replace one input matrix with its low-rank approximation
using truncated SVD. These algorithms are suitable for inference where the
weight matrix factorization can be pre-computed offline, but are not applicable
to training since the high cost of factorization is incurred in every matrix product.

\noindent{\bf Column-row sampling
(CRS)~ \citep{drineas2001fast,drineas2006fast}} 
The sampling algorithm approximates matrix product $A^\top B$ by sampling $k$ columns of $A^\top$ and
respective rows of $B$ to form smaller matrices,
which are then multiplied as usual. 

We choose CRS as the basis for our current work because it meets all the criteria above: 
It is applicable to fully-connected layers of any size, 
its effectiveness does not depend on the matrix contents, 
its sampling is computationally lightweight, and may use regular matrix multiplication 
algorithms since the sampled sub-matrices remain dense.  

\subsection{CRS}
Let $A\in\mathbb{R}^{n\times m}, B\in\mathbb{R}^{n\times p}$. 
Their product $A^\top B$ is approximated as a weighted sum of  
outer products between sampled columns of $A^\top$ and corresponding rows of
$B$:
\begin{align}
A^\top B\approx\sum_{t=1}^{k}\frac{1}{kp_{i_{t}}}A^{\top(i_{t})} B_{(i_{t})}
\label{eq:approx}
\end{align}
where $A^{\top(i)},B_{(i)}$ denote the matrix $i$'th column and row respectively, $k$ is the number of samples (satisfying $1\leq k\leq n$), 
$\{p_i\}_{i=1}^{n}$ is a probability distribution over the column-row pairs of $A^\top ,B$ and $i_{t}\in\left\{1,...,n\right\}$. This algorithm allows
linear reduction in complexity from $O(mnp)$ to	$O(mkp)$.
	
~\eqref{eq:approx} can also be expressed 
as $A^\top DS^\top SDB$, where $D\in\mathbb{R}^{n\times n}$ is a diagonal scaling matrix with:
\begin{align}
    (D)_{j,j}=\frac{1}{\sqrt{kp_j}}
\end{align}

and $S\in\mathbb{R}^{k\times n}$ is a sampling matrix that selects $k$ features, possibly with replacement. $S$ is a random matrix, where each row has 1 in one entry and zeros in others. In each row, the probability of having the non-zero entry in column $j$ is $p_j$.

\citet{drineas2006fast} show that CRS is unbiased:
\begin{align}
\E{A^\top DS^\top SDB} = A^\top B    
\end{align}
and also derive upper bounds for the expected Frobenius and spectral norms of the error matrix $\norm{A^\top B-A^\top DS^\top SDB}$. They show that the error is minimized when the sampling probabilities are proportional to the product of the column-row Euclidean norms:

\begin{align}
\label{eq_optimal_prob}
p_{i}=\frac{\abs{A_{(i)}}\abs{B_{(i)}}}{\sum_{j=1}^{n}\abs{A_{(j)}}\abs{ B_{(j)}}}
\end{align}

In which case the expected Frobenius error is:
\begin{align}
\frac{1}{k}\left(\sum_{t=1}^k\abs{A_{(i_t)}}\abs{B_{(i_t)}}\right)^2-\frac{1}{k}\norm{A^\top B}_F^2
\label{eq:CRS_frobenius_error}
\end{align}

\subsection{Approximate Tensor Operations and Dropout}
Sampling-based approximations can be interpreted as a flavor of Dropout~ \citep{srivastava2014dropout}, a popular technique to prevent overfitting by randomly zeroing individual activations during training. However, the sparsity pattern resulting from Dropout is unstructured and therefore cannot be exploited efficiently by GPUs despite recent advances in structured sparsity support\citep{mishra2021accelerating}. Prior works on fast Dropout training \citep{wang2013fast, graham2015efficient} are different than ours and do not demonstrate acceleration of large networks while maintaining accuracy. 

Some works proposed non-uniform Dropout probabilities for individual activations \citep{li2016improved} or full channels \citep{hou2019weighted}. Their sampling heuristics are different from ours which are derived from optimal approximation. Furthermore, they use Dropout only for preventing overfitting and do not leverage it to speed up training. In our experiments we demonstrate the utility of sampling-based approximations for DNNs with and without Dropout. Conversely, we did not observe improved accuracy from approximations which could have been attributed to reduced overfitting.

\section{Approximate Linear Regression}
\label{section:approximate_linear_regression}
We now analyze CRS in the simpler context of linear regression, where we can derive the effects of approximations in closed form. We show that this leads to biased gradient estimates.

Let $X\subset\mathbb{R}^{n\times M}$ a dataset containing $M$ examples, each a vector in $\mathbb{R}^n$.
Every $x^i\in X$ is associated with a "ground truth" value $y^i\in\mathbb{R}$.

Let $w\in\mathbb{R}^n$ be parameters in a linear model that predicts a value $\bar{y}^i\in\mathbb{R}$ for every $x^i\in X$:
\begin{align}
\bar{y}^i = w^\top x^i
\end{align}

To simplify the notation we do not include an explicit bias term. We do so without loss of generality since we can always add another entry of $1$ to the features. 

Let us define the MSE (Mean Square Error) loss function:
\begin{align}
\ell = \sum_{i=1}^M(\bar{y}^i-y^i)^2
\end{align}

When using SGD (Stochastic Gradient Descent), we are given a single example $x^i\in\mathbb{R}^n$ in each step, and update $w$ using the gradients $\frac{\partial \ell}{\partial w}$. For notation simplicity we omit the superscript $i$ from $x^i,\bar{y}^i,y^i$.

The gradients are given by the chain rule as:
\begin{align}
\frac{\partial \ell}{\partial w} &= 2x(w^\top x-y)
\label{eq:liner_regression_dw}
\end{align}

Now, let us assume that the multiplication $w^\top x$ is approximated using CRS. The linear regression model now becomes:
\begin{align}
\hat{y}=w^\top DS^\top SDx
\label{eq:liner_regression_crs_y}
\end{align}

and for the MSE loss the gradients will be:
\begin{align}
\widehat{\frac{\partial \ell}{\partial w}} = 2DS^\top SDx(w^\top DS^\top SDx-y)
\label{eq:liner_regression_approx_dw}
\end{align}

Where $\widehat{\frac{\partial \ell}{\partial w}}$ denotes the CRS weight gradients.

Note that $DS^\top SD$ appears twice in~\eqref{eq:liner_regression_approx_dw}. In $w^\top DS^\top SDx$ it represents sampling in the forward pass, while in $DS^\top SDx$ it results in passing gradients only to the elements of $w$ that were sampled in the forward pass.

It should be emphasized that \eqref{eq:liner_regression_crs_y} and \eqref{eq:liner_regression_approx_dw} in fact describe gradients with respect to a different loss function compared to \eqref{eq:liner_regression_dw}: one loss function uses $\hat{y}$ while the other uses $\bar{y}$. If the approximate gradients are unbiased estimates of the non-approximate gradients, we could relate the approximate training process to the original one. However, the weight gradients do not satisfy this unbiasedness property:
\begin{align}
\E{(\widehat{\frac{\partial \ell}{\partial w}})_j}=2x_j\left(\sum_{t=1}^nw_t\E{(\tilde{S})_{j,j}(\tilde{S})_{t,t}}x_t-y\right)
\label{eq:liner_regression_approx_dw_element}
\end{align}
where we denote $\tilde{S}\triangleq DS^\top SD$, and use the fact that $\tilde{S}$ is diagonal.

For the expression in~\eqref{eq:liner_regression_approx_dw_element} to be equal to that in~\eqref{eq:liner_regression_dw} we need that $\E{(\tilde{S})_{j,j}(\tilde{S})_{t,t}}=1$. However, this is not the case because $(\tilde{S})_{j,j}$ and $(\tilde{S})_{t,t}$ are not independent random variables: an entry in the diagonal of $\tilde{S}$ is the (scaled) number of times a column-row pair was selected out of the $k$ total samples in CRS. The event of selecting a particular pair therefore affects selecting others.

We note that if instead of changing the loss function we treated the approximate multiplication as a "black box" that replaces the original product, we could use~\eqref{eq:liner_regression_dw} and only replace the forward pass product $w^\top x$ with the "black box" substitute of $w^\top DS^\top SDx$. This would yield:
\begin{align}
\widetilde{\frac{\partial \ell}{\partial w}} = 2x(w^\top DS^\top SDx-y)
\label{eq:liner_regression_approx_dw_blackbox}    
\end{align}
which satisfies:
\begin{align}
\E{\widetilde{\frac{\partial \ell}{\partial w}}}=\frac{\partial \ell}{\partial w}.    
\end{align}
~\eqref{eq:liner_regression_approx_dw_blackbox} is equivalent to applying the approximate computation in the forward pass, but propagating the gradients to all weight entries in the same way as if the computation were exact. In practice we find that this approach leads to significantly lower accuracy in deep neural networks compared to sampling the same entries in the forward and backward pass, or to applying approximations in the backward pass only.

\section{Bernoulli-CRS}
\label{section:bernoulli_sampling}
We now turn to develop Bernoulli-CRS, a new variant of sampling approximation that enables to sample column-row pairs independently and without replacement. Applied to linear regression, we show that using Bernoulli-CRS is equivalent to employing unbiased gradient estimates in addition to a bias term which can be interpreted as scale dependent weight regularization.

\textbf{Bernoulli-CRS:}
These aforementioned properties can be achieved by assigning a separate Bernoulli sampling probability $p_i$ for each column-row pair $i$, and sampling pairs independently of each other. To control the amount of sampling, we add another constraint that all the probabilities will sum up to an integer $k$:
\begin{align}
\sum_{i=1}^n p_i = k
\end{align}
Let us define $K\in\mathbb{R}^{n \times n}$ a random diagonal sampling matrix where $K_{j,j}\sim \text{Bernoulli}(p_j)$ for $1 \leq j \leq n$. Furthermore, let us define another diagonal scaling matrix $P\in\mathbb{R}^{n\times n}$ where $P_{j,j}=\frac{1}{\sqrt{p_j}}$ for $1 \leq j \leq n$.

Using the $K$ and $P$ matrices we may now define our new Bernoulli-CRS algorithm. Let $A\in\mathbb{R}^{n\times m}$ and $B\in\mathbb{R}^{n\times p}$. The product $A^\top B$ can be approximated with $\tilde{A}^\top \tilde{B}$  defined as follows:
\begin{align}
\tilde{A}^\top \tilde{B}: = \sum_{i=1}^{n}\frac{Z_i}{p_{i}}A^{\top(i)} B_{(i)}
= A^\top PKKPB
\end{align}
where $\{ Z_i\sim \text{Bernoulli}(p_i) \}_{i=1}^n$ are independent random variables.
We denote $\tilde{A}\triangleq KPA$ and $\tilde{B}\triangleq KPB$.

In the appendix we develop the properties of Bernoulli-CRS. We show it is unbiased and derive bounds on the error variance both in expectation and in high probability. We derive the optimal sampling probabilities minimizing the expected variance, and show that under certain conditions they are given by the simpler expression:
\begin{align}
p_i=\min\left\{\frac{k\abs{A_{(i)}}\abs{B_{(i)}}}{\sum_{j=1}^n\abs{A_{(j)}}\abs{B_{(j)}}},1\right\}
\end{align}

In the appendix we show that applying Bernoulli-CRS in linear regression leads to unbiased estimate of the original gradients with an additional regularization term $\mathcal{R}(w)$, which we define as:

\begin{align}
\mathcal{R}(w) 
&
= \mathbf{E}\left[ \sum_{j=1}^n\frac{1-p_j}{p_j}x_j^2 w_j^2\right]
\end{align}
and the expectation is with respect to the distribution of the data samples.

The term $\mathcal{R}(w)$ can be interpreted as input-dependent $L_2$ regularization. The regularization is higher as $x_j$ grows in magnitude and as $p_j$ decreases. Both serve to reduce the impact on the weights if they were chosen with small probabilities or mostly because of the input size. We note that \citet{wager2013dropout} conduct a similar analysis for Dropout in the particular case where the same sampling probability is used for all features. 

To summarize, sampling in the simpler case of linear regression minimizes the original loss function with an added regularization term.

\section{Approximate Backpropagation in Non-Linear Deep Networks}
\label{section:backprop}
The analysis of approximate linear regression cannot simply generalize to deep non-linear networks: non-linearity leads to biased network output even if the approximate multiplication is itself unbiased. Still, we are able to obtain strong theoretical results on the relations between exact and approximate training if the approximations are limited to the backward pass: the forward pass is calculated as usual, and the matrix products in the backward pass are performed using approximate matrix multiplication. 

We prove the following theorem:

\begin{theorem}
Let $f(x,W,b)$ be a multi-layer neural network with $\beta$-Lipschitz activation functions $\sigma$. Let $\ell$ be a $\beta$-Lipschitz loss function, and let the network be trained with SGD using properly decreasing learning rate. Assume that the weights are bounded; and further assume that the matrix products in the backward pass are approximated using an unbiased approximation scheme, i.e., 
$$
\E{A^\top B-\mathtt{approx}(A^\top B)}=0
$$
and that there exists a constant $C$ and a norm $\norm{\cdot}$ such that:
$$
\E{\norm{A^\top B-\mathtt{approx}(A^\top B)}^2}\leq C\norm{A}^2\norm{B}^2.
$$
Then the approximated NN gradient estimates are unbiased, and their second moments are bounded.
\label{thm:sgd}
\end{theorem}
\begin{*corollary}
Based on recent works on non-convex optimization (see e.g.~ \cite{ge2015escaping}), the unbiasedness and bounded second moments ensured by Theorem~\ref{thm:sgd} imply that approximate backpropagation enjoys the same convergence guarantees as regular SGD training.
\end{*corollary}

In the appendix we show that CRS and other sampling algorithms satisfy the property
$$
\E{\norm{A^\top B-\mathtt{approx}(A^\top B)}^2}\leq C\norm{A}^2\norm{B}^2
$$

Note that for Theorem~\ref{thm:sgd} we required that weights will be bounded during the training process. This is a strong assumption which could be justified if weight regularization or clipping is used. In the appendix we prove the same results without relying on these assumptions, if only the weight gradients are approximated and if the activation function is bounded (such as sigmoid). 

\section{Sampling Without Scaling and Top-$k$ Selection}
\label{section:topk}
We now consider a different sampling scheme where $k$ column-row pairs are selected deterministically without scaling. This can be viewed as a special case of Bernoulli-CRS, where the sampling probabilities are either 0 or 1. We now show that under certain assumptions on the distribution of the input matrices, this scheme can lead to the optimal estimation:

\begin{theorem}
Let $A$ be an $n\times m$ random matrix and $B$ be an $n\times p$ random matrix, such that 
$$
\E{A^{\top(i)} B_{(i)}}=0
$$ for $1\leq i \leq n$.
Assume $k$ column-row pairs with indices $\{j\}_1^n$ are sampled from $A$ and $B$. Then, the MMSE estimator of the product $A^\top B$ is $\tilde{A}^\top \tilde{B}$ where $\tilde{A},\tilde{B}$ are constructed from the sampled column-row pairs without scaling.

Furthermore, if $A^{\top(i)} B_{(i)}$ and $A^{\top(j)} B_{(j)}$
are independent for different $i,j$ then the MSE is minimized when sampling $k$ pairs with the maximum norm multiplication $\abs{A_{(i)}}\abs{B_{(i)}}$.
\label{thm:topk}
\end{theorem}

The assumptions in Theorem~\ref{thm:topk} can hold in practice if weights are initialized with a zero-centered distribution\citep{glorot2010understanding}, if the distribution of weights remains centered around zero during training \citep{glorot2010understanding, blundell2015weight, han2015learning, thoma2017analysis}, and if different weights can be considered pairwise-independent \citep{huang2020convolution}.

\begin{figure}[h]
  \subfigure[Matrix product: both matrix entries drawn from \protect$\mathcal{N}(1,1)$]{\includegraphics[width=0.45\linewidth]{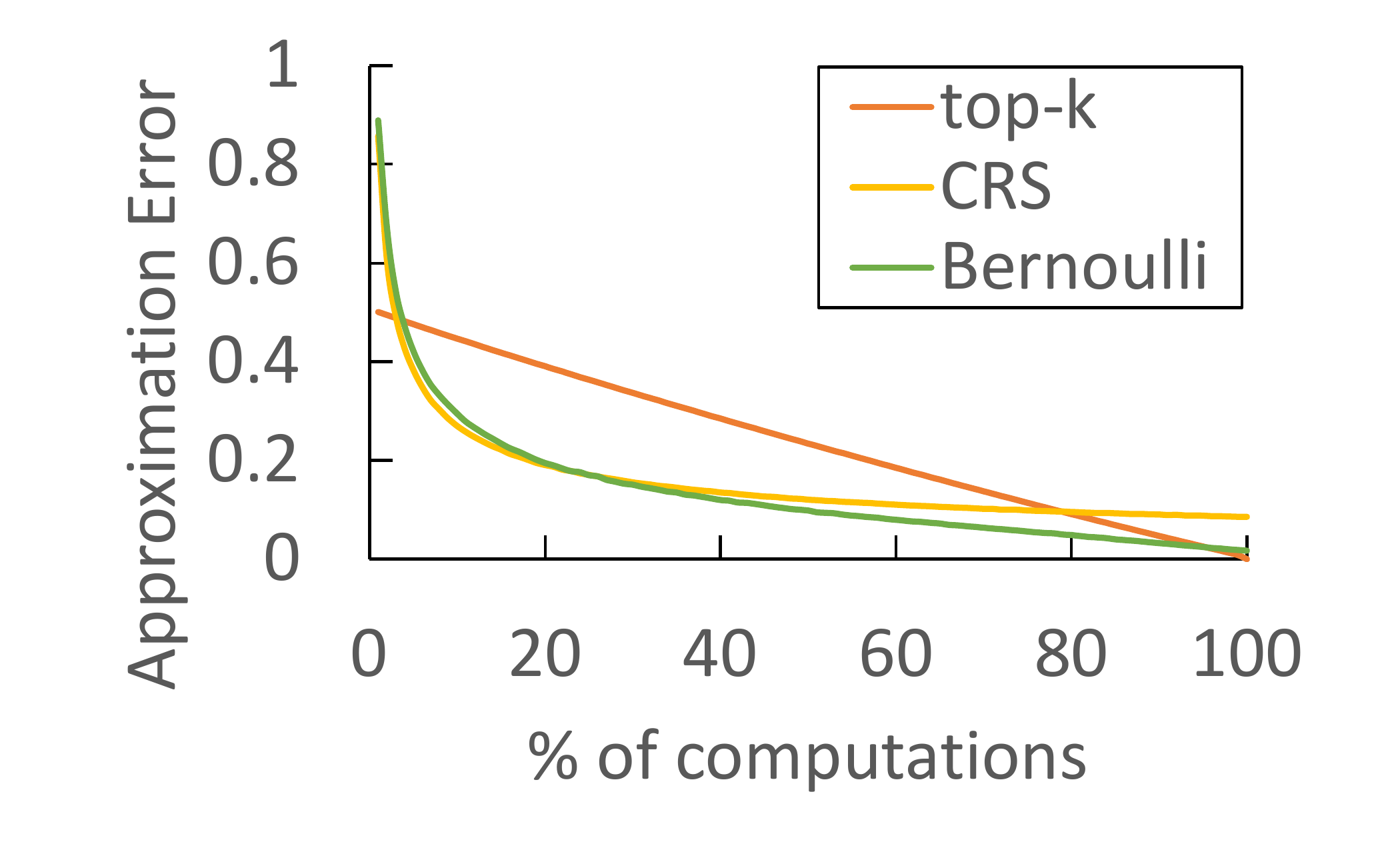}
  \label{fig:normal_1_1_100x100_frobenius}}\hfill~
  \subfigure[Matrix product: one matrix entries drawn from \protect$\mathcal{N}(0,1)$, the other from \protect$\mathcal{N}(1,1)$]{\includegraphics[width=0.45\linewidth]{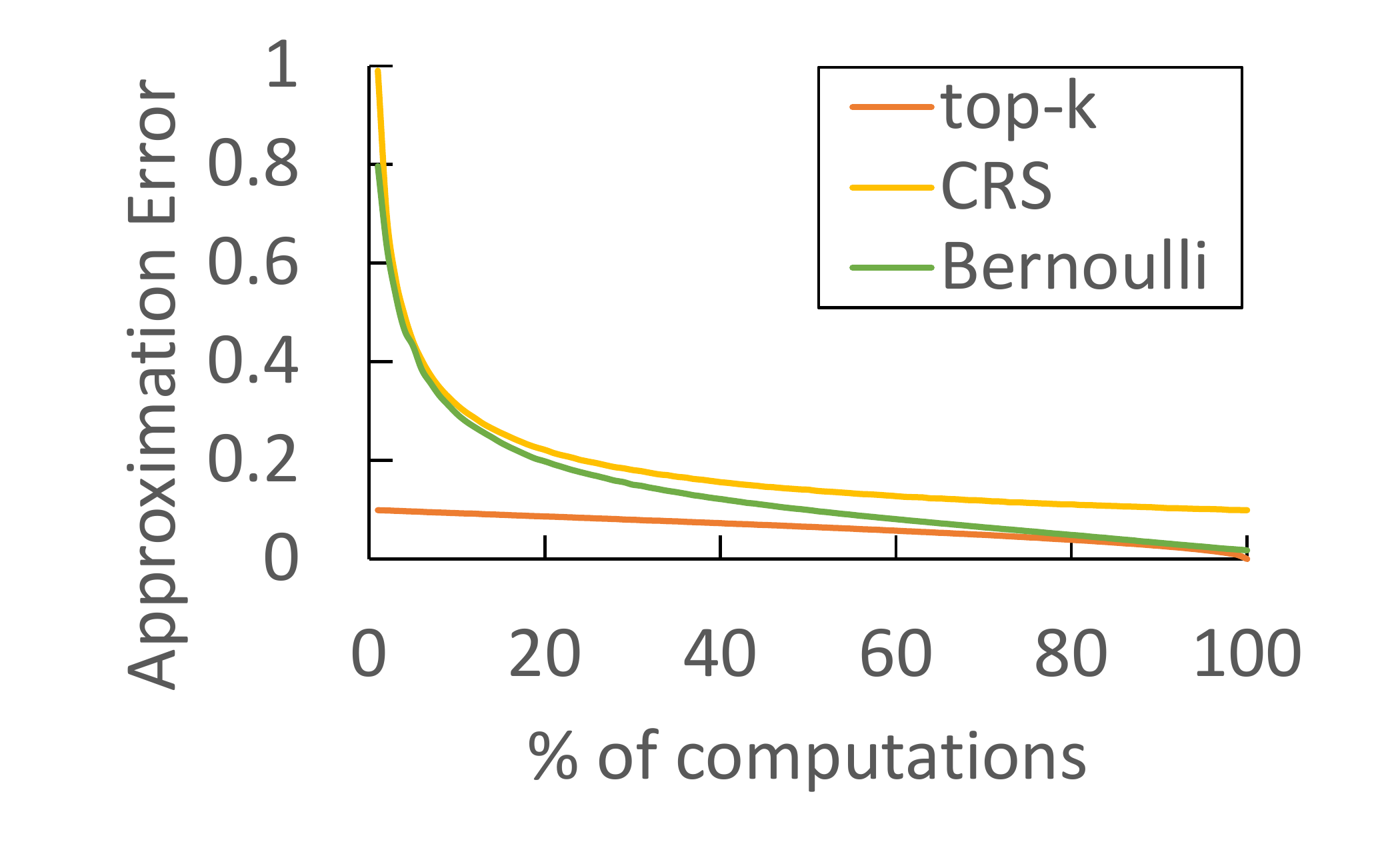}
  \label{fig:normal_0_1_1_1_100x100_frobenius}}
  \caption{Approximation error depending on the amount of performed computations. Lower is better.}
  \label{fig_synthetic}
\end{figure}

We study the approximation quality of CRS, Bernoulli-CRS and top-$k$ selection on synthetic matrix multiplication.
We generate $100\times 100$ random matrices and compute the error metric:
\begin{align}
\frac{\norm{A^\top B-\mathtt{approx}(A^\top B)}_F}{\norm{A}_F\norm{B}_F} \end{align}
Figures~\ref{fig:normal_1_1_100x100_frobenius},\ref{fig:normal_0_1_1_1_100x100_frobenius}
show the approximation error for different algorithms and sampling ratios, averaged over 1000 runs. We observe that Bernoulli-CRS outperforms CRS in higher sampling ratios. Also, when one matrix has i.i.d entries with zero mean, Bernoulli-CRS outperforms CRS and top-$k$ selection performs the best as expected from Theorem \ref{thm:topk}. 

We also consider a different flavor of top-$k$ selection, which we refer to as "top-$k$-weights": sampling $k$ column-row pairs corresponding to rows of $B$ with the highest norms. While not providing  the theoretical guarantees of Theorem~\ref{thm_sup:topk}, the new variant has a desirable property for data parallel distributed training, where weights are identical between different workers. A deterministic selection algorithm that only depends on the weights will sample the same weights for all workers, allowing to reduce the gradient communication between nodes to the selected weights only.

\section{Approximating Convolutions}
\label{section:conv}
\begin{figure}[h]
  \includegraphics[width=1\linewidth]{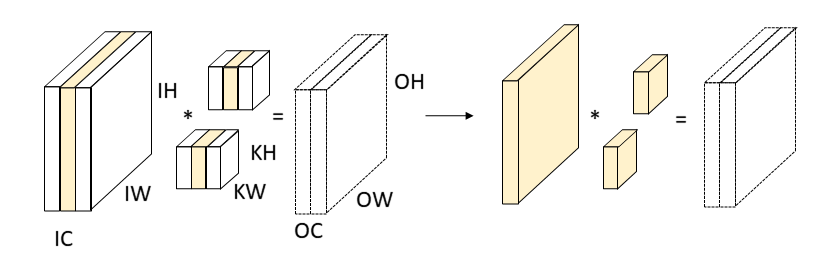}
    \caption{Sampling one input channel out of three}
  \label{fig:sample_conv}
\end{figure}

We extend the basic CRS algorithm to the approximation of multi-channel convolutions. In matrix multiplication sampling is performed over the common dimension. The analogue for multi-channel convolution is to sample over the input channels dimension, illustrated in Figure~\ref{fig:sample_conv}. As in the matrix case, the output dimensions remain the same. 

In the appendix we derive the optimal sampling probabilities and scaling factors. Bernoulli-CRS and top-$k$ algorithms can be developed for convolutions as well in an analogous way.

\section{Experimental Results}
We implement CRS, Bernoulli-CRS and top-$k$ selection approximation algorithms in PyTorch both for matrix multiplication and convolution. Our implementation allows to control the sampling degree and the application of approximation in the forward or backward passes.

We replace exact tensor operations with their sampling-based approximations, \emph{without changing training hyper-parameters}. Only column-row pairs sampled in the forward pass are used during backpropagation as the rest do not affect the loss. Hence, sampling in the forward pass also reduces the amount of backward pass computations by the same ratio. We apply approximations only during training, and use exact computations for validation/test evaluation.

We evaluate our approximate training technique on several network architectures and datasets: MLP and CNN on MNIST, Wide ResNet 28-10 on CIFAR-10 and ResNet-50 and ResNet-152 on ImageNet. We train the networks on a single node using NVidia V100 GPUs (two GPUs for ResNet-152, one for the rest), and measure the reduction in multiply-accumulate operations due to sampling as well as the overall speedup in total training time versus the non-approximate baseline. The appendix includes additional details on the models and the training process. 

\begin{table*}[t]
\caption{Compute reduction, communication reduction and wall-clock speedup of training with approximate tensor operations.}
\label{sample-table}
\vskip 0.15in
\begin{center}
\begin{small}
\begin{sc}
\begin{tabular}{lccccc}
\toprule
Network     & & Compute & Communication& Accuracy    & Training\\
            & & Reduction  & Reduction & (Baseline)   & Speedup\\
\midrule
\multicolumn{2}{l}{MLP (MNIST)} & 50\%  & - & 98.22\% (98.22\%) & -  \\
\midrule
\multicolumn{2}{l}{CNN (MNIST)} & 66\%  & - & 99.25\% (99.35\%) & -  \\
\midrule
\multicolumn{2}{l}{WRN-28-10 (CIFAR-10)}  & 50\% & - & 95.89\% (96.17\%) &  1.33x    \\ 
\midrule
\multicolumn{2}{l}{ResNet-50 (ImageNet)}  & 6.5\% &- & 75.63\% (75.6\%)  &  1.04x   \\ 
\midrule
\multirow{5}{*}{ResNet-152}&\multirow{2}*{Single Node}
            & 40\%   & - & 76.44\% (77.65\%)   &  1.16x  \\\cmidrule{3-6}
             & & 9\%    & - & 77.66\% (77.65\%)   & 1.04x  \\\cmidrule[2pt]{2-6}
            (ImageNet) &\multirow{3}*{8 Nodes} & 40\%   & 48\% & 76.44\% (77.65\%) & 1.37x \\\cmidrule{3-6}
            & & 12\%   & 23\% & 77.48\% (77.65\%) & 1.13x \\\cmidrule{3-6}
            & & 9\%   & 13\% & 77.8\% (77.65\%) & 1.09x \\
\bottomrule
\end{tabular}
\end{sc}
\end{small}
\end{center}
\vskip -0.1in
\label{table_summary}
\end{table*}

\begin{figure*}[h]
\centering
  \begin{minipage}{.32\textwidth}
  \includegraphics[width=\linewidth]{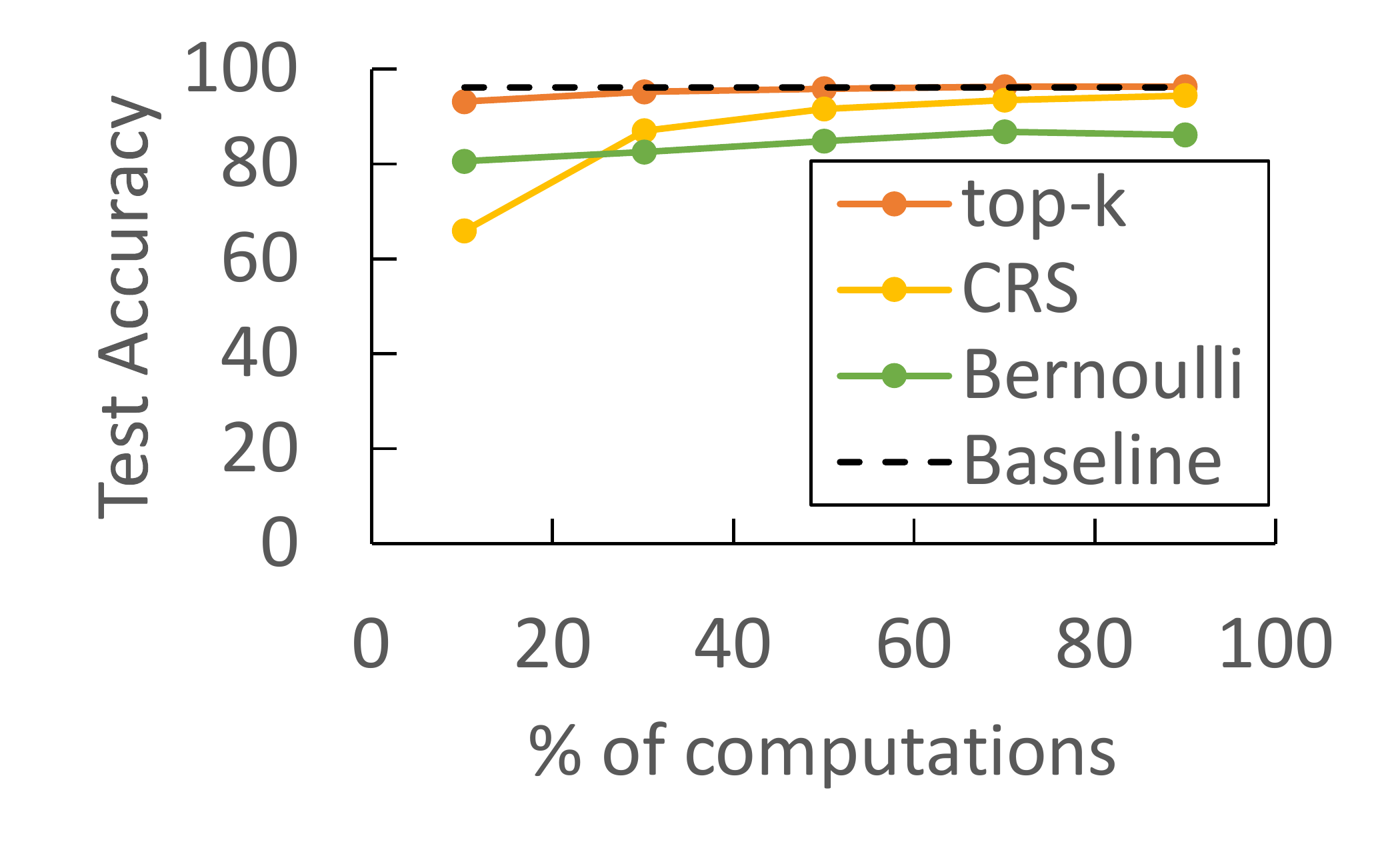}
  \caption{WRN-28-10 on CIFAR-10 (approximate forward and backward)}
  \label{fig:wrn_cifar10_forward}
  \end{minipage}
  ~
  \begin{minipage}{.32\textwidth}
  \includegraphics[width=\linewidth]{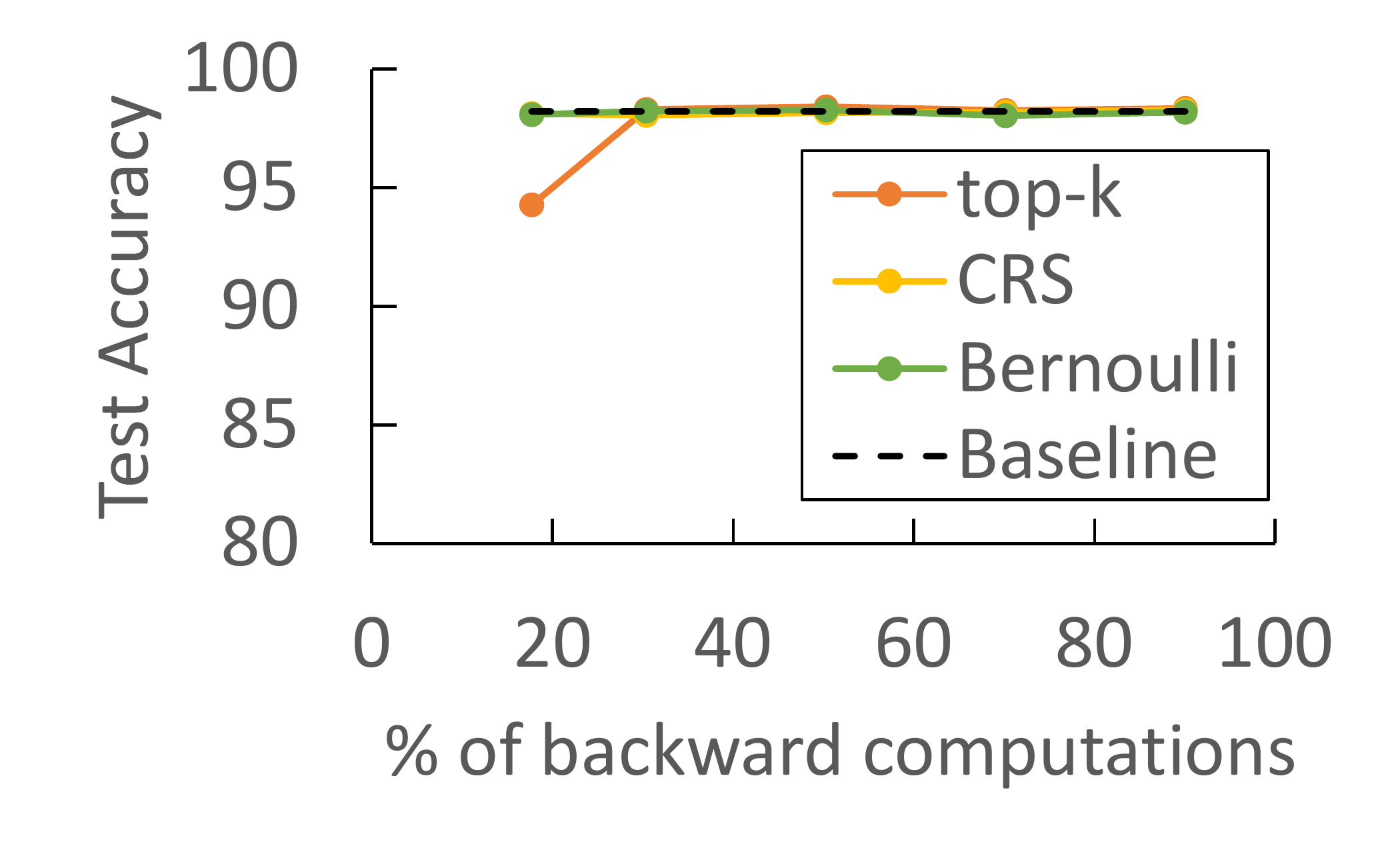}
  \caption{3-layer MLP on MNIST (exact forward, approximate  backward)}
  \label{fig:mnist_mlp_backward}
  \end{minipage}
  ~
  \begin{minipage}{.32\textwidth}
  \includegraphics[width=\linewidth]{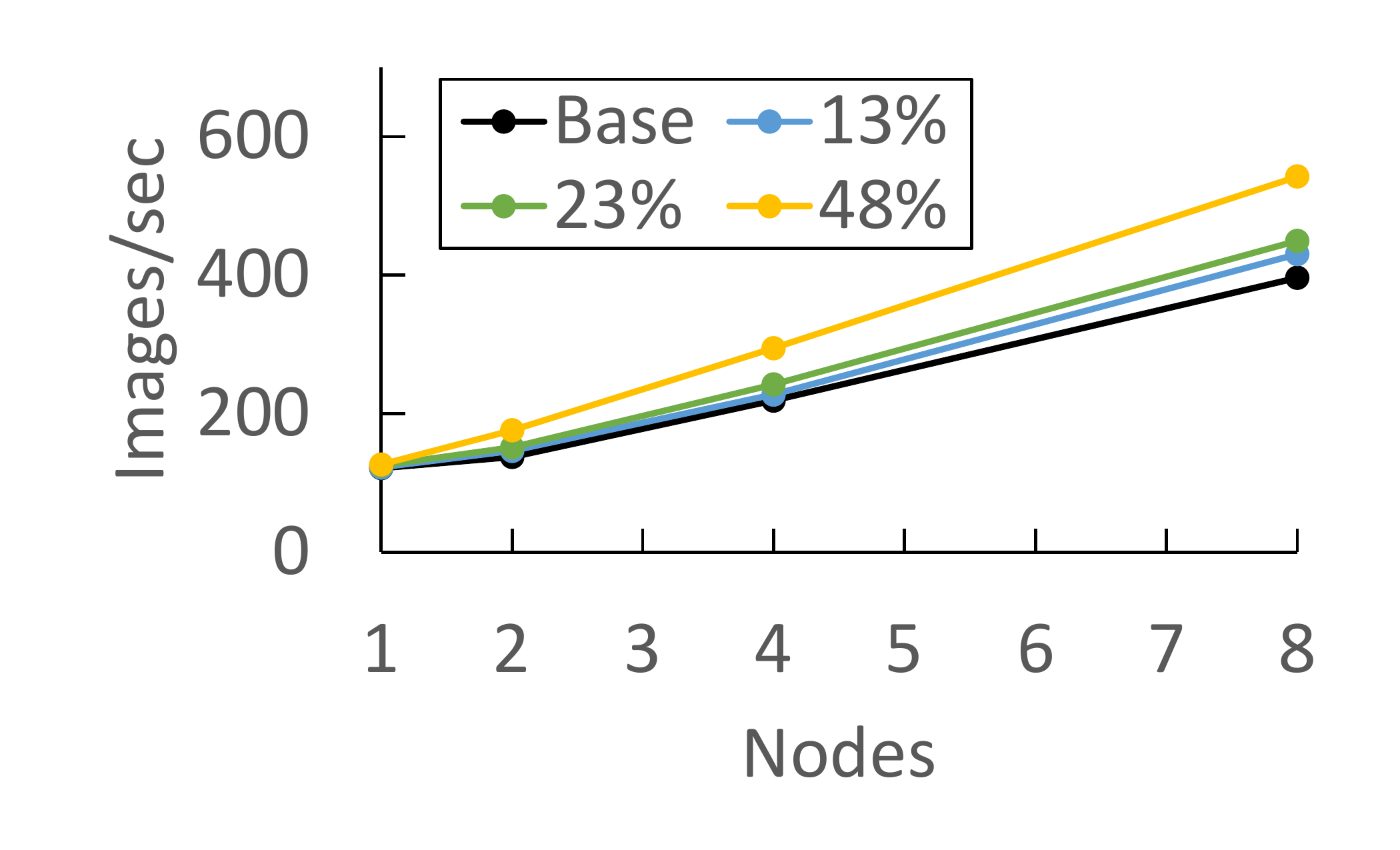}
  \caption{AllReduce with top-k-weights sampling (\% fewer gradients sent).}
  \label{fig:dist_compcomm}
  \end{minipage}
\end{figure*}

Our results using top-$k$ sampling are summarized in Table~\ref{table_summary}. We see a reduction of up to 66\% in the computational cost with
little or no degradation in model accuracy, and up to 1.33x faster training time. We believe the gap between compute reduction and end-to-end speedup can be reduced by fusing the sampling with the matrix multiplication routines, or running on a different HW architecture that allows fast sampling and norm computation. We note that the small MNIST models do not exhibit training time speedup since they are not compute-intensive enough to saturate the GPU. The ratio between compute reduction and actual speedup is smaller in ResNet-152 compared to ResNet-50 and WRN-28-10 because the batch size per GPU is lower due to the limited GPU memory capacity. 

\paragraph{Sampling Algorithms}
We compare CRS, Bernoulli-CRS and top-$k$ selection on MNIST and CIFAR-10 and find empirically that top-$k$ results in higher accuracy and faster training time (Figure~\ref{fig:wrn_cifar10_forward}). This result is consistent with that of approximate $\mathcal{N}(0,1)$ matrix product (Figure~\ref{fig:normal_0_1_1_1_100x100_frobenius}). This is not surprising given Theorem~\ref{thm_sup:topk} and our empirical observation that the weight distribution is close to symmetrical around zero throughout training.

\paragraph{Approximations in Forward Pass and Backpropagation} For the small MNIST models we are able to perform as low as 10\% of the computations in the backward pass without harming accuracy (Fig.~\ref{fig:mnist_mlp_backward}). However, in the larger models (WRN-28-10) we find empirically that accuracy drops when approximating only the backward pass. Therefore, in Table~\ref{table_summary} we report results for consistent sampling in the forward and backward passes.  

\begin{figure}[h]
  \subfigure[WRN-28-10 CIFAR-10]{\includegraphics[width=0.32\linewidth]{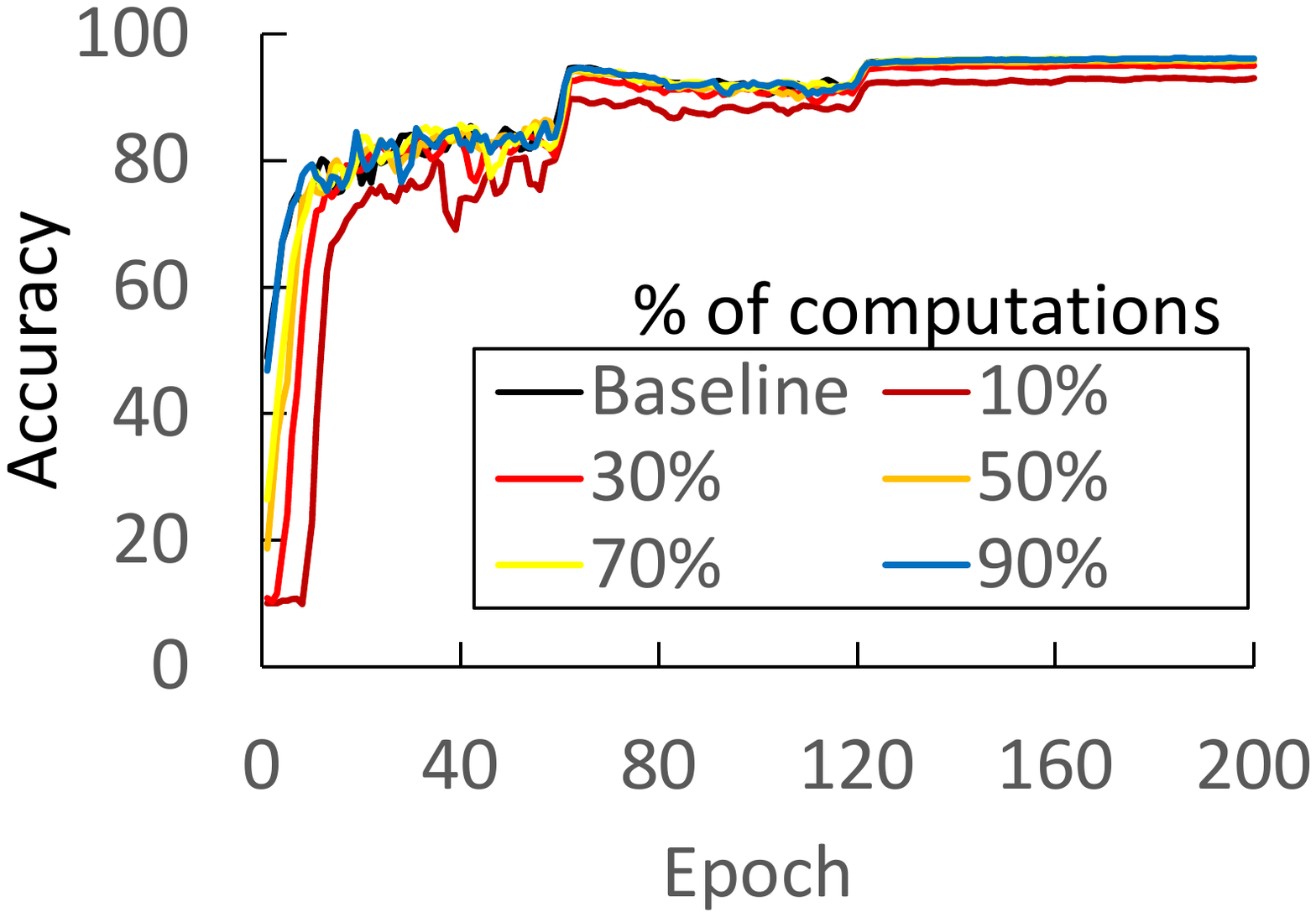}
  \label{fig:wrn_cifar10_learning}}~
  \subfigure[ResNet-50 Imagenet (top-1)]{\includegraphics[width=0.32\linewidth]{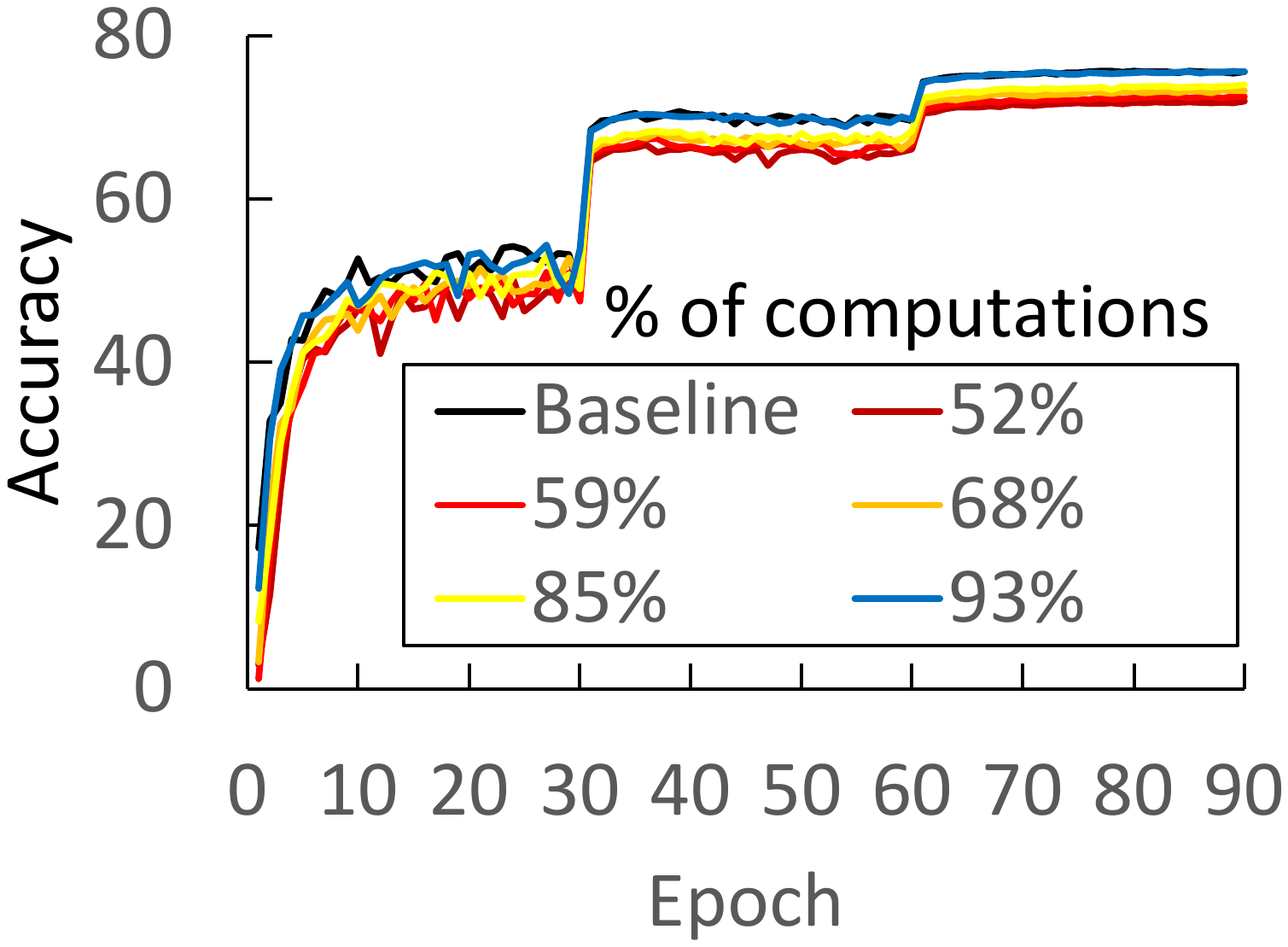}
  \label{fig:resnet50_learning}}~
  \subfigure[ResNet-152 Imagenet (top-1)]{\includegraphics[width=0.32\linewidth]{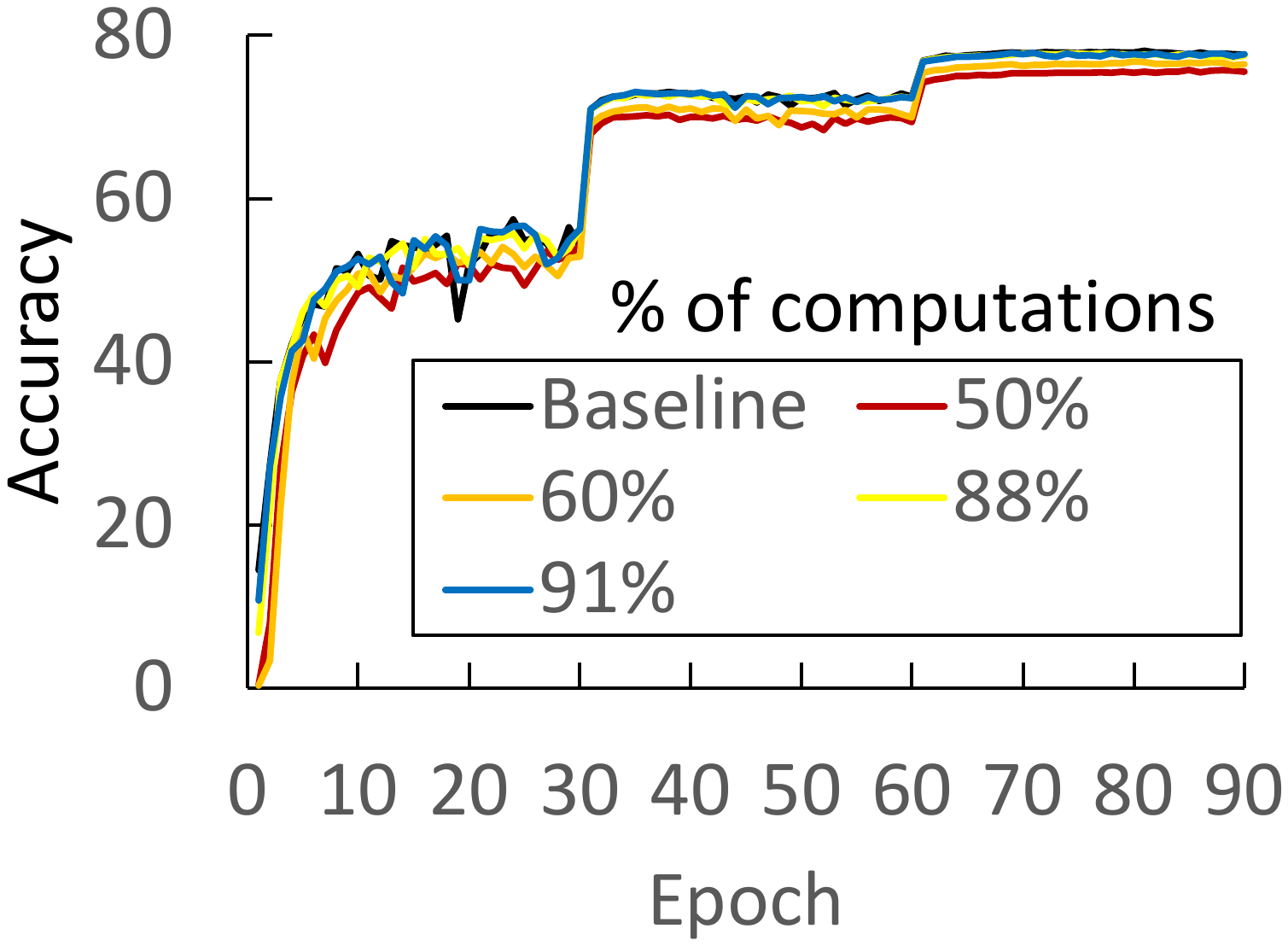}
  \label{fig:resnet152_learning}}
  \caption{Learning curves for validation accuracy under different top-$k$ sampling ratios}
  \label{fig:learning_curves}
\end{figure}

\paragraph{Sampling Ratio Sensitivity} We find that the achievable compute reduction is not consistent across networks and datasets. For MNIST and CIFAR-10 we maintain good accuracy while reducing 50\%-66\% of the computations. However, ImageNet proved to be more sensitive and we kept the accuracy intact when applying 50\% sampling to the ResNet layers with 1024 or more channels only. Figure~\ref{fig:learning_curves} shows the learning curves under different sampling ratios compared to the non-approximate baseline.

\paragraph{Distributed Training} We implement an AllReduce "top-$k$-weights" scheme in PyTorch. This scheme performs reduction only for the gradients of the sampled weights, reducing inter-node communications. Table~\ref{table_summary} shows the accuracy-speedup trade-off for ResNet-152 distributed training. 
Figure~\ref{fig:dist_compcomm} shows the respective scaling behavior of these schemes relative to the exact baseline. We note that compute savings did not lead to significant single-node speedup since in this experiment the V100 GPUs (from Amazon AWS) had lower memory capacity, which led to smaller batch size per GPU. The multi-node training speedup is therefore mostly due to the communication savings.
\section{Conclusion}
In this work we have demonstrated the utility of sample-based approximation of tensor operations for neural network training, both theoretically and empirically. We believe that further acceleration could be achieved through dedicated GPU primitives fusing sampling and matrix multiplication/convolution, as well as varying and adaptive sampling rates for different layers and iterations. Studying other approximation algorithms, applications in resource-constrained environments and bridging the gaps between our theoretical results and what worked best in practice are all promising directions for future research. Overall, we believe that sample-based approximations and fast approximations in general are valuable additions to the toolbox of techniques for deep learning acceleration.

\section*{Acknowledgments and Disclosure of Funding}
K.Y. Levy acknowledges support from the Israel Science Foundation (grant No. 447/20). I. Hakimi acknowledges support from the Hasso Plattner Institute at the Technion. M. Silberstein acknowledges support from the Israel Science Foundation (grant No.  1027/18).

\begin{small}
\bibliographystyle{unsrtnat}
\bibliography{neurips_2021}
\end{small}

\appendix
\newpage
\section*{Appendices}
\section{Bernoulli-CRS Properties}
\label{section:bernoulli_sampling_sup}
Let us define $K\in\mathbb{R}^{n \times n}$ a random diagonal sampling matrix where $K_{j,j}\sim \text{Bernoulli}(p_j)$ for $1 \leq j \leq n$.

Let us define another diagonal scaling matrix $P\in\mathbb{R}^{n\times n}$ where $P_{j,j}=\frac{1}{\sqrt{p_j}}$ for $1 \leq j \leq n$.

Using the $K$ and $P$ matrices we may now define our new Bernoulli-CRS algorithm. Let $A\in\mathbb{R}^{n\times m}$ and $B\in\mathbb{R}^{n\times p}$. The product $A^\top B$ can be approximated with $\tilde{A}^\top \tilde{B}$  defined as follows:
\begin{align}
\tilde{A}^\top \tilde{B}: = \sum_{i=1}^{k}\frac{Z_i}{p_{i}}A^{\top(i)} B_{(i)}
= A^\top PKKPB
\end{align}
where $\{ Z_i\sim \text{Bernoulli}(p_i) \}_{i=1}^n$ are independent random variables.
We denote $\tilde{A}\triangleq KPA$ and $\tilde{B}\triangleq KPB$.

First, we show that the above holds in expectation:

\begin{proposition}
$\E{\tilde{A}^\top \tilde{B}}=A^\top B$.
\label{proposition:expectation}
\end{proposition}

Let $T=\text{Trace}(K)$ the number of non-zero diagonal elements in $K$. We note that to perform the actual computation it is enough to sample the $T$ column-row pair with the corresponding element in $K$ being non-zero. Unlike CRS, the lower rank of the sampled matrices is not constant and depends on the random matrix $K$. Its expectation is controlled through the parameter $k$:
\begin{proposition}
$\E{T}=k$.
\end{proposition}
Therefore, Bernoulli-CRS will perform on average the same amount of computations as in the fixed-rank CRS.

Let us further derive the properties of the proposed sampling algorithm. Specifically, what are the optimal values for the probabilities $p_i$ under the constraint $\sum_{i=1}^n p_i = k$?

First, let us calculate the variance of $\tilde{A}^\top \tilde{B}$:
\begin{proposition}
$$
\Var{(\tilde{A}^\top \tilde{B})_{i,j}}=\sum_{t=1}^n\frac{1-p_t}{p_t}A_{t,i}^2B_{t,j}^2
$$
\label{proposition:variance}
\end{proposition} 

We will be interested in the Frobenius norm of the error matrix $\norm{A^\top B-\tilde{A}^\top \tilde{B}}_F^2$, which can be derived from the following theorem:
\begin{theorem}
The expected Frobenius norm of the error matrix $\E{\norm{A^\top B-\tilde{A}^\top \tilde{B}}_F^2}$ is $\sum_{t=1}^n\frac{1-p_t}{p_t}\abs{A_{(t)}}^2\abs{B_{(t)}}^2$.

Furthermore, under the constraint $\sum_{i=1}^n p_i = k$ it is minimized for the probabilities:

$$
p_i = \frac{\abs{A_{(i)}}\abs{B_{(i)}}}{\sqrt{\mu}}1_{\{0<\abs{A_{(i)}}\abs{B_{(i)}}<\mu \}} + 1_{\{\abs{A_{(i)}}\abs{B_{(i)}}\geq \mu \}}
$$

where $\mu$ is a root of the following function:

$$
G(\mu): = \sum_{i=1}^n \left(\frac{\abs{A_{(i)}}\abs{B_{(i)}}}{\sqrt{\mu}}1_{\{0<\abs{A_{(i)}}\abs{B_{(i)}}<\mu \}} + 1_{\{\abs{A_{(i)}}\abs{B_{(i)}}\geq \mu \}} \right) -k
$$
\label{thm:expected_frobenius}
\end{theorem}
\begin{*corollary}
The sampling probabilities
\begin{align*}
p_i=\min\left\{\frac{k\abs{A_{(i)}}\abs{B_{(i)}}}{\sum_{j=1}^n\abs{A_{(j)}}\abs{B_{(j)}}},1\right\}
\end{align*}
are optimal if $k\leq \frac{\sum_{i=1}^n\abs{A_{(i)}}\abs{B_{(i)}}}{\max_i\abs{A_{(i)}}\abs{B_{(i)}}}$
\end{*corollary}

From Theorem~\ref{thm:expected_frobenius} it follows that for the  probabilities:
\begin{align}
p_i=\min\left\{\frac{k\abs{A_{(i)}}\abs{B_{(i)}}}{\sum_{j=1}^n\abs{A_{(j)}}\abs{B_{(j)}}},1\right\}
\end{align}
the expected Frobenius error is:
\begin{align}
\frac{1}{k}\left(\sum_{i=1}^n e_i\abs{A_{(i)}}\abs{B_{(i)}}\right)^2-\sum_{i=1}^n e_i\abs{A_{(i)}}^2\abs{B_{(i)}}^2
\end{align}
where we denote: 
\begin{align}
e_i \triangleq
\begin{cases} 
      1 & \abs{A_{(i)}}\abs{B_{(i)}} \leq\frac{\sum_{j=1}^n \abs{A_{(j)}}\abs{B_{(j)}}}{k} \\
      0 & \text{else}
   \end{cases}.
\end{align}

Comparing that with the bound in~\eqref{eq:CRS_frobenius_error}, we can see that different values of $A,B$ determine which algorithm performs better.

Knowing the expected Frobenius error also implies a bound on the spectral norm of the error matrix, since the spectral and Frobenius norms are related by:
\begin{align}
\norm{A}\leq\norm{A}_F\leq\sqrt{r}\norm{A}
\end{align}
where $r$ is the rank of $A$ and $\norm{A}$ denotes its spectral norm.

The following theorem yields high probability bounds for the Frobenius and spectral norms for the Bernoulli-CRS algorithm:
\begin{theorem}
Let $A\in\mathbb{R}^{n\times m}$ and $B\in\mathbb{R}^{n\times p}$. Let $\tilde{A},\tilde{B}$ be the sampled matrices according to the Bernoulli-CRS algorithm described above.
Denote
$$
R \triangleq \max_i\norm{A^{\top(i)} B_{(i)}}
$$
and
$$
\sigma^2 \triangleq \frac{1}{k}\left(\sum_{i=1}^n e_i\abs{A_{(i)}}\abs{B_{(i)}}\right)^2-\sum_{i=1}^n e_i\abs{A_{(i)}}^2\abs{B_{(i)}}^2
$$
then, for all $t\geq 0$:
$$
\mathbb{P}\left\{\norm{A^\top B - \tilde{A}^\top \tilde{B}}\geq t\right\} \leq (m+p)\cdot \exp\left(\frac{-t^2/2}{\sigma^2+Rt/3}\right)
$$
$$
\mathbb{P}\left\{\norm{A^\top B - \tilde{A}^\top \tilde{B}}_F\geq t\right\} \leq (m+p)^{3/2}\cdot \exp\left(\frac{-t^2/2}{\sigma^2+Rt/3}\right)
$$
\label{thm:bernoulli_berenstein}
\end{theorem}

\subsection{Bernoulli-CRS in linear regression}
We now show that applying Bernoulli-CRS in linear regression leads to unbiased estimate of the original gradients with an additional term that can be interpreted as regularization.
The analysis for linear regression using Bernoulli-CRS is the same as in Section~\ref{section:approximate_linear_regression}, with the sampling and scaling matrices $DS^\top SD$ replaced with $PKKP$

The expression for the weight gradient (simimlar to~\eqref{eq:liner_regression_approx_dw_element}) now becomes:
\begin{align}
& \E{(\widehat{\frac{\partial \ell}{\partial w}})_j}=2x_j(\sum_{t=1}^nw_t(\E{(\tilde{K})_{j,j}(\tilde{K})_{t,t}}x_t-y) \\
&= 2x_j(w^\top x-y + w_j\left(\E{(\tilde{K})_{j,j}^2}-1\right)x_j) \\
&= 2x_j(w^\top x-y + \frac{1-p_j}{p_j}w_jx_j)
\label{eq:liner_regression_bernoulli_dw_element}
\end{align}
where we denote $\tilde{K}\triangleq PKKP$.

When comparing \eqref{eq:liner_regression_bernoulli_dw_element} and \eqref{eq:liner_regression_dw} we see that using Bernoulli-CRS yields unbiased estimates of the original gradients with an additional bias term that is related to a scale-dependent regularization $\mathcal{R}(w)$, which we define as:
\begin{align}
\mathcal{R}(w) 
&
= \mathbf{E}\left[ \sum_{j=1}^n\frac{1-p_j}{p_j}x_j^2 w_j^2\right]
\end{align}
and the expectation is with respect to the distribution of the data samples.

This term can be interpreted as input-dependent $L_2$ regularization. The regularization is higher as $x_j$ grows in magnitude and as $p_j$ decreases. Both serve to reduce the impact on the weights if they were chosen with small probabilities or mostly because of the input size.

\section{Approximating Convolutions - Details}
\label{section:conv_details_sup}
Formally, let $I\in\mathbb{R}^{IW\times IC}_{B\times IH}$ be the input tensor, where $B$ is the batch size, $IH,IW$ are the input height and width, and $IC$ are the input channels. Let $K\in\mathbb{R}_{KH\times KW}^{IC\times OC}$ be the kernels tensor, where $KH,KW$ are the kernel height and width, and $IC,OC$ are the input and output channels respectively. Let $O\in\mathbb{R}^{OW\times OC}_{B\times OH}$ be the output tensor, where $OH,OW$ are the output height and width.

The multi-channel convolution operation is defined as:
\begin{align}
\label{eq_convolution}
O_{b,oh}^{ow,oc}=I*K = \sum_{i=1}^{IC}\sum_{kh=1}^{KH}\sum_{kw=1}^{KW}I_{b,oh+kh-1}^{ow+kw-1,i}\cdot K_{kh,kw}^{i,oc}
\end{align}

For notation simplicity, we assume zero padding. The inner sums in~\eqref{eq_convolution} can be written as 1-channel convolutions: 

\begin{align}
O_{b,oh}^{ow,oc}= \sum_{i=1}^{IC}I^{[i]}*K_{[i]}
\end{align}

where $I^{[i]}\in\mathbb{R}^{IW\times 1}_{B\times IH}$ denotes a 
tensor with one input channel that corresponds to the $i$'th input channel 
of $I$, i.e ${I^{[i]}}_{b,ih}^{iw,1}=I_{b,ih}^{iw,i}$. Similarly, 
$K_{[i]}\in\mathbb{R}^{1\times OC}_{KH\times KW}$ corresponds to the 
$i$'th input channel of $K$.

This formulation immediately hints at the possibility to sample over the input channel dimension, similarly to sampling column-row pairs in matrices.  
We propose to approximate convolutions by sampling lower-rank tensors:
\begin{align}
\tilde{O}=\sum_{t=1}^{k}\frac{1}{kp_{i_t}}I^{[i_{t}]}*K_{[i_{t}]}\triangleq\tilde{I}*\tilde{K}
\label{eq_approximate_convolution_tensor}
\end{align}
where $\{i_t\}_{t=1}^{k}$ are such that $i_t\in\{1,...,IC\}$ and $\{p_i\}_{i=1}^{IC}$ is a probability distribution over the input channels, $\tilde{I}$ is a tensor composed of sampled input channels of $I$ scaled by	$\sqrt{\frac{1}{kp_{i}}}$, and $\tilde{K}$ is a tensor composed of corresponding sampled input channels of $K$ scaled by the same factor.

Computing the convolution of the smaller tensors $\tilde{I}*\tilde{K}$ can be done using standard efficient convolution implementations. Figure~\ref{fig:sample_conv} illustrates the sampling operation.

The properties of the approximation in~\eqref{eq_approximate_convolution_tensor} can be derived similarly 
to the CRS derivations for matrix multiplication. 
In particular, we prove the approximation is unbiased, and similar to matrix CRS, we use sampling probabilities proportional to the tensor Euclidean norms: 
\begin{align}
p_i=\frac{\big\Vert I^{[i]}\big\Vert_{F}\cdot\big\Vert 
K_{[i]}\big\Vert_{F}}{\sum_{j=1}^{IC}
\big\Vert I^{[j]}\big\Vert_{F}\cdot\big\Vert K_{[j]}\big
\Vert_{F}}
\label{eq_conv_prob}
\end{align}

In section \ref{section:conv_proofs} we show that the optimal sampling probabilities are significantly more complicated to calculate, but under certain conditions they reduce to~\eqref{eq_conv_prob}.

Bernoulli-CRS and top-$k$ algorithms can be developed for convolutions as well in an analogous way.

\section{Proofs}
\label{section:proofs}

\subsection{Proofs for Section~\ref{section:backprop} - Approximate Backpropagation}
\begin{theorem_sup}
Let $f(x,W,b)$ be a multi-layer neural network with $\beta$-Lipschitz activation functions $\sigma$. Let $\ell$ be a $\beta$-Lipschitz loss function, and let the network be trained with SGD using properly decreasing learning rate. If the matrix products in the backward pass are approximated using an unbiased approximation scheme satisfying: 
$$
\E{A^\top B-\mathtt{approx}(A^\top B)}=0
$$
and:
$$
\E{\norm{A^\top B-\mathtt{approx}(A^\top B)}^2}\leq C\norm{A}^2\norm{B}^2
$$
for some finite constant $C$ and some norm $\norm{\cdot}$,

and if the weights are bounded, then the approximated gradients are unbiased with bounded second moments.  
\label{thm_sup:sgd}
\end{theorem_sup}

\begin{*corollary}
Based on recent works on non-convex optimization \cite{ge2015escaping}, Theorem~\ref{thm:sgd} implies that approximate backpropagation enjoys the same convergence guarantees as regular SGD training.
\end{*corollary}

\begin{proof}
The network $f$ can be described by:
\begin{align*}
h_1 &= W_1^\top x+b_1 \\
a_1 &= \sigma(h_1) \\
h_l &= W_l^\top a_{l-1}+b_l \\
a_l &= \sigma(h_l) \\
\hat{y} &= W_L^\top a_{L-1}
\end{align*}
where $x\in\mathbb{R}^n$,$W_1\in\mathbb{R}^{n\times{d_1}}$, $W_l\in\mathbb{R}^{d_{l-1}\times{d_l}}$, $b_l\in\mathbb{R}^{d_l}$, $\ell$ is the number of layers and $\hat{y}\in\mathbb{R}^{d_L}$ is the network output. 

Let us denote the weight, bias and activation gradients with respect to a loss function $\ell$ by $\nabla W_l,\nabla b_l,\nabla a_l$ respectively. Let us denote and the gradients yielded by the approximation scheme as $\nabla\tilde{W}_l,\nabla\tilde{b}_l,\nabla\tilde{a}_l$.

\begin{lemma_sup}
$$
\E{\nabla\tilde{W}_l}=\nabla W_l\quad\text{and}\quad\E{\nabla\tilde{b}_l}=\nabla b_l
$$
\label{lemma_sup:unbiased_sgd}
\end{lemma_sup}
\begin{proof}
We prove by induction. The last layer satisfies:
\begin{align*}
\nabla W_L = a_{L-1}\nabla \hat{y} \quad\quad\quad
\nabla a_{L-1} = W_L\nabla \hat{y}
\end{align*}
and its approximation is given by:
\begin{align*}
\nabla\tilde{W}_L = \mathtt{approx}(a_{L-1}\nabla \hat{y}) \quad\quad\quad
\nabla\tilde{a}_{L-1} = \mathtt{approx}(W_L\nabla \hat{y})
\end{align*}
Since the approximation methods satisfies:
$$
\E{A^\top B-\mathtt{approx}(A^\top B)}=0
$$
we get:
\begin{align*}
\E{\nabla\tilde{W}_L} = \nabla W_L \quad\quad\quad
\E{\nabla\tilde{a}_{L-1}} = \nabla a_{L-1}
\end{align*}
for the induction step, we will show that if $\E{\nabla\tilde{a}_l} = \nabla a_l$ then:
\begin{align*}
\E{\nabla\tilde{W}_{l-1}} &= \nabla W_{l-1} \\
\E{\nabla\tilde{b}_{l-1}} &= \nabla b_{l-1} \\
\E{\nabla\tilde{a}_{l-1}} &= \nabla a_{l-1}
\end{align*}
$\nabla\tilde{W}_{l-1}$ is given by:
$$
\nabla\tilde{W}_{l-1} = \mathtt{approx}(a_{l-1}\nabla \tilde{h}_l^\top ) = \mathtt{approx}(a_{l-1}\Sigma'(h_l)\nabla \tilde{a}_l^\top )
$$
where $\Sigma'(h_l)$ is a diagonal matrix with the diagonal being the derivative of $\sigma$ in location $h_l$.
Taking the expectation we get:
\begin{align*}
\E{\nabla\tilde{W}_{l-1}} &= \E{\mathtt{approx}(a_{l-1}\Sigma'(h_l)\nabla \tilde{a}_l^\top )} \\
&= \E{\E{\mathtt{approx}(a_{l-1}\Sigma'(h_l)\nabla \tilde{a}_l^\top )|\nabla\tilde{a}_l^\top }} \\
&= \E{a_{l-1}\Sigma'(h_l)\nabla \tilde{a}_l^\top } \\
&= a_{l-1}\Sigma'(h_l)\nabla a_l^\top \\
&= \nabla W_{l=1}
\end{align*}
where we used the unbiased approximation property of \texttt{approx} and the law of total expectation.
Similar arguments for $\E{\nabla\tilde{a}_{l-1}}$ yield:
\begin{align*}
\E{\nabla\tilde{a}_{l-1}} &= \E{\mathtt{approx}(W_l\nabla\tilde{h}_l)} \\
&= \E{\mathtt{approx}(W_l\Sigma'(h_l)\nabla\tilde{a}_l)} \\
&= \E{\E{\mathtt{approx}(W_l\Sigma'(h_l)\nabla\tilde{a}_l)|\nabla\tilde{a}_l}} \\
&= \E{W_l\Sigma'(h_l)\nabla\tilde{a}_l} \\
&= W_l\Sigma'(h_l)\nabla a_l \\
&= \nabla a_{l-1}
\end{align*}
and for $\E{\nabla\tilde{b}_{l-1}}$:
\begin{align*}
\E{\nabla\tilde{b}_{l-1}} &= \E{\nabla\tilde{h}_{l-1}} \\
&= \E{\Sigma'(h_l)\nabla\tilde{a}_l} \\
&= \Sigma'(h_l)\nabla a_l \\
&= \nabla b_{l-1}
\end{align*}

\end{proof}
In other words, the unbiased estimation of the gradients follows from the linearity of backpropagation with respect to the gradients, even for non-linear activation functions. 

We can write the training step using SGD and the approximate gradients $\nabla\tilde{W}_l^t$ for layer $l$ at iteration $t$ as:
$$
W_{l}^{t+1} = W_l^t -\alpha_t(\nabla W_l^t+\omega_t)
$$
where $\omega_t$ is a gradient noise defined as:
$$
\omega_t \triangleq \nabla\tilde{W}_l^t-\nabla W_l^t
$$

Based on Lemma~\ref{lemma_sup:unbiased_sgd}, the gradient noise $\omega_t$ is a martingale difference sequence satisfying:
$$
\E{\omega_t|W_{t-1}}=\E{\nabla\tilde{W}_l^t-\nabla W_l^t|W_{t-1}}=0
$$

\begin{lemma_sup}
Under the assmuptions in Theorem~\ref{thm_sup:sgd}:
$$
\E{\norm{\omega_t}^2|W_{t-1}}<D
$$
for some constant $D$.
\label{lemma_sup:bounded_2nd_momemt_sgd}
\end{lemma_sup}
\begin{proof}
We prove by induction. Since $\ell$ is $\beta$-Lipschitz, the gradients $\nabla y$ are bounded. During backpropagation the gradients are propagated by:
$$
\nabla\tilde{a}_{l-1} = \mathtt{approx}(W_l\Sigma'(h_l)\nabla\tilde{a}_l)
$$
Let us assume $\nabla\tilde{a}_l$ is bounded and show that $\nabla\tilde{a}_{l-1}$ is bounded in expectation as well: 

\begin{align*}
\E{\norm{\nabla\tilde a_{l-1}}^2} &\leq \E{\norm{\nabla\tilde a_{l-1}-W_l\Sigma'(h_l)\nabla\tilde{a}_l}^2} + \E{\norm{W_l\Sigma'(h_l)\nabla\tilde{a}_l}^2} \\
&\leq C\norm{W_l}^2\norm{\Sigma'(h_l)\nabla\tilde{a}_l}^2 \\
& < D'
\end{align*}

for some constant $D'$, where the second inequality follows from the properties of \texttt{approx} and last inequality follows from the $\beta$-Lipschitz of $\Sigma$, the induction assumption on the boundness of $\nabla\tilde{a}_l$ and the assumption on the boundness of $W_l$. 

The gradients $\nabla\tilde{W}$ are calculated by:
$$
\nabla\tilde{W}_{l-1} = \mathtt{approx}(a_{l-1}\Sigma'(h_l)\nabla\tilde{a}_l^\top )
$$
and therefore:

\begin{align*}
\E{\norm{\omega_t}^2|W_{t-1}} &= \E{\norm{\nabla\tilde{W}_l^t-\nabla W_l^t}^2|W_{t-1}} \\
&= \E{\norm{\nabla\tilde{W}_l^t-(a_{l-1}\Sigma'(h_l)\nabla\tilde{a}_l^\top )+(a_{l-1}\Sigma'(h_l)\nabla\tilde{a}_l^\top )+  \nabla W_l^t}^2|W_{t-1}} \\
&\leq \E{\norm{\nabla\tilde{W}_l^t-a_{l-1}\Sigma'(h_l)\nabla\tilde{a}_l^\top }^2}+\E{\norm{a_{l-1}\Sigma'(h_l)\nabla\tilde{a}_l^\top }^2} + \E{\norm{\nabla W_l^t}^2} \\
&\leq C_1\norm{a_{l-1}}^2\E{\norm{\Sigma'(h_l)\nabla\tilde{a}_l^\top }^2} + C_2\norm{a_{l-1}}^2\E{\norm{\Sigma'(h_l)\nabla\tilde{a}_l^\top }^2}\ + \E{\norm{\nabla W_l^t}^2} \\
&\leq D
\end{align*}
In the second inequality we used the properties of $\texttt{approx}$. In the last inequality we used the boundness of $\Sigma',\nabla W_l^t$ from the assumptions, the boundness of $\E{\norm{\nabla\tilde{a}_l^\top }^2}$ from above. In addition, we assumed boundness of the activations $a_{l-1}$. This assumption holds if the activation function $\sigma$ is bounded (for example, sigmoid) and in the general case it also requires the assumptions on the boundness of weights and inputs. 
\end{proof}

The same arguments can be made for the bias and the approximate bias gradients.

Based on Lemmas~\ref{lemma_sup:unbiased_sgd} and \ref{lemma_sup:bounded_2nd_momemt_sgd} and using standard analysis of SGD (for example \cite{bertsekas1996neuro} and \cite{ge2015escaping}) the SGD convergence guarantees hold for approximate backpropagation as well.
\end{proof}

\begin{remark}
Both CRS and Bernoulli-CRS satisfy the property
$$
\E{\norm{A^\top B-\mathtt{approx}(A^\top B)}^2}\leq C\norm{A}^2\norm{B}^2
$$
since the expected Frobenius norm for the error matrix satisfies:
\begin{align*}
&\E{\norm{A^\top B-\tilde{A}^\top \tilde{B}}_F^2} = \\
&\frac{1}{k}\left(\sum_{i=1}^n e_i\abs{A_{(i)}}\abs{B_{(i)}}\right)^2-\sum_{i=1}^n \abs{A_{(i)}}^2\abs{B_{(i)}}^2 \\
&\leq \left(\sum_{i=1}^n e_i\abs{A_{(i)}}\abs{B_{(i)}}\right)^2 \\
&\leq \left(\sum_{i=1}^n \abs{A_{(i)}}^2\right)\left(\sum_{i=1}^n \abs{B_{(i)}}^2\right) \\
&= \norm{A}_F^2\norm{B}_F^2
\end{align*}
where we used Theorem~\ref{thm_sup:expected_frobenius} and the Cauchy-Schwarz inequality.
\end{remark}

\begin{*corollary}
Let $f(x,W,b)$ be a multi-layer neural network with bounded $\beta$-Lipschitz activation functions $\sigma$. Let $\ell$ be a $\beta$-Lipschitz loss function, and let the network be trained with SGD using properly decreasing learning rate. If the weight gradient matrix products in the backward pass are approximated using an unbiased approximation scheme satisfying: 
$$
\E{A^\top B-\mathtt{approx}(A^\top B)}=0
$$
and:
$$
\E{\norm{A^\top B-\mathtt{approx}(A^\top B)}^2}\leq C\norm{A}^2\norm{B}^2
$$
for some finite constant $C$ and some norm $\norm{\cdot}$,

then then the approximated gradients are unbiased with bounded second moments.   
\label{thm_sup:sgd_without_bounded_weights}
\end{*corollary}
\begin{proof}
Lemma~\ref{lemma_sup:unbiased_sgd} under these assumptions holds by the same arguments.
We now prove the equivalent of Lemma~\ref{lemma_sup:bounded_2nd_momemt_sgd}:
\begin{align*}
\E{\norm{\omega_t}_F^2|W_{t-1}} &= \E{\norm{\nabla\tilde{W}_l^t-\nabla W_l^t}_F^2|W_{t-1}} \\
&= \E{\norm{\nabla\tilde{W}_l^t-a_{l-1}\Sigma'(h_l)\nabla a_l^\top }_F^2} \\
&\leq \norm{a_{l-1}}_F^2\norm{\Sigma'(h_l)\nabla a_l^t}_F^2 \\
& \leq D
\end{align*}
The first inequality follows from the properties of $\texttt{approx}$. The second inequality follows from the $\beta$-Lipschitz property of $\ell,\Sigma$ bounding the second term, and from the boundness of the activation function $\sigma$ bounding the first term.
\end{proof}

\subsection{Proofs for Section~\ref{section:topk} - Sampling Without Scaling and Top-$k$ Selection}
\begin{theorem_sup}
Let $A$ be a $n\times m$ random matrix and $B$ be $n\times p$ random matrix, such that 
$$
\E{A{^\top(i)} B_{(i)}}=0
$$ for $1\leq i \leq n$.
Assume $k$ column-row pairs with indices $\{j\}_1^n$ are sampled from $A$ and $B$.

Then, the MMSE estimator for the matrix product $A^\top B$ would be $\tilde{A}^\top \tilde{B}$ where $\tilde{A},\tilde{B}$ are constructed from the sampled column-row pairs without scaling.

Furthermore, if $A^{\top(i)} B_{(i)}$ and $A^{\top(j)}, B_{(j)}$
are independent for different $i$ and $j$ then the MSE will be minimized when sampling $k$ pairs with the maximum norm multiplication $\abs{A_{(i)}}\abs{B_{(i)}}$.
\label{thm_sup:topk}
\end{theorem_sup}
\begin{proof}
Given sampled pairs $j_1,...,j_k$ the MMSE estimator would be:
\begin{align*}
\widehat{A^\top B} &= \E{A^\top B|A_{(j_1)},...,A_{(j_k)},B_{(j_1)},...,B_{(j_k)}} \\
&=\E{\sum_{i=1}^kA_{(j_i)}^\top B_{(j_i)}+\sum_{i\notin \{j\}_1^k}A^{\top(i)} B_{(i)}|A_{(j_1)},...,A_{(j_k)},B_{(j_1)},...,B_{(j_k)}} \\
&= \sum_{i=1}^kA_{(j_i)}^\top B_{(j_i)}+\sum_{i\notin \{j\}_1^k}\E{A^{\top(i)} B_{(i)}} \\
&= \sum_{i=1}^kA_{(j_i)}^\top B_{(j_i)} \\
&= \tilde{A}^\top \tilde{B}
\end{align*}

The MSE would be:
\begin{align*}
\E{\norm{A^\top B-\tilde{A}^\top \tilde{B}}_F^2} = \E{\norm{\sum_{i\notin \{j\}_1^k}A^{\top(i)} B_{(i)}}_F^2}
\end{align*}

if we assume independence between different column-row pairs $A^{\top(i)} B_{(i)},A^{\top(j)} B_{(j)}$ then the last expression reduces to:
$$
\sum_{i\notin \{j\}_1^k}\E{\norm{A^{\top(i)} B_{(i)}}_F^2}=\sum_{i\notin \{j\}_1^k}\E{\abs{A_{(i)}}^2\abs{B_{(i)}}^2}
$$
and therefore will be minimized for a top-$k$ selection scheme that samples the pairs with the highest norm.
\end{proof}

\subsection{Proofs for Section~\ref{section:bernoulli_sampling_sup} - Bernoulli-CRS}
\begin{proposition_sup}
$\E{\tilde{A}^\top \tilde{B}}=A^\top B$
\label{proposition_sup:expectation}
\end{proposition_sup}
\begin{proof}
\begin{align*}
\E{A^\top PKKPB} &= A^\top PP\E{KK}B \\
 &= A^\top PP\E{K}B \\
 &= A^\top B
\end{align*}
where we used that fact that $K$ is diagonal and that $K_{i,i}\in\{0,1\}$.
\end{proof}

\begin{proposition_sup}
$\E{T}=k$
\end{proposition_sup}
\begin{proof}
$$
\E{T} = \E{\sum_{j=1}^nK_{j,j}}=\sum_{j=1}^n\E{K_{j,j}}=\sum_{j=1}^np_j=k
$$
\end{proof}

\begin{proposition_sup}
$$
\Var{(\tilde{A}^\top \tilde{B})_{i,j}}=\sum_{t=1}^n\frac{1-p_t}{p_t}A_{t,i}^2B_{t,j}^2
$$
\label{proposition_sup:variance}
\end{proposition_sup} 
\begin{proof}
Fix $i,j$. From Proposition~\ref{proposition_sup:expectation}:
$$
\E{(\tilde{A}^\top \tilde{B})_{i,j}} = (A^\top B)_{i,j}
$$  

Calculating the second moment:
\begin{align*}
\E{(\tilde{A}^\top \tilde{B})_{i,j}^2} &=\E{\left(\sum_{t=1}^nA_{t,i}\frac{K_{t,t}}{p_t}B_{t,j}\right)^2}\\
&=\E{\sum_{t=1}^n\sum_{u=1}^n A_{t,i}\frac{K_{t,t}}{p_t}B_{t,j}A_{u,i}\frac{K_{u,u}}{p_u}B_{u,j}} \\
&=\E{\sum_{t=1}^n\sum_{u\neq t}^n A_{t,i}\frac{K_{t,t}}{p_t}B_{t,j}A_{u,i}\frac{K_{u,u}}{p_u}B_{u,j}} \\
&+ \E{\sum_{t=1}^nA_{t,i}^2B_{t,j}^2\frac{K_{t,t}}{p_t^2}} \\
&= \sum_{t=1}^k\sum_{u\neq t}^k A_{t,i}B_{t,j}A_{u,i}B_{u,j} + \sum_{t=1}^n\frac{1}{p_t}A_{t,i}^2B_{t,j}^2 \\
&= (A^\top B)_{i,j}^2-\sum_{t=1}^nA_{t,i}^2B_{t,j}^2+ \sum_{t=1}^n\frac{1}{p_t}A_{t,i}^2B_{t,j}^2 \\
&= (A^\top B)_{i,j}^2 + \sum_{t=1}^n\frac{1-p_t}{p_t}A_{t,i}^2B_{t,j}^2 
\end{align*}
Therefore:
\begin{align*}
\Var{(\tilde{A}^\top \tilde{B})_{i,j}} &= \E{(\tilde{A}^\top \tilde{B})_{i,j}^2}-\E{(\tilde{A}^\top \tilde{B})_{i,j}}^2 \\
& = \sum_{t=1}^k\frac{1-p_t}{p_t}A_{t,i}^2B_{t,j}^2
\end{align*}
\end{proof}

\begin{theorem_sup}
The expected Frobenius norm of the error matrix $\E{\norm{A^\top B-\tilde{A}^\top \tilde{B}}_F^2}$ is $\sum_{t=1}^n\frac{1-p_t}{p_t}\abs{A_{(t)}}^2\abs{B_{(t)}}^2$.

Furthermore, under the constraint $\sum_{i=1}^n p_i = k$ it is minimized for the probabilities:

$$
p_i = \frac{\abs{A_{(i)}}\abs{B_{(i)}}}{\sqrt{\mu}}1_{\{0<\abs{A_{(i)}}\abs{B_{(i)}}<\mu \}} + 1_{\{\abs{A_{(i)}}\abs{B_{(i)}}\geq \mu \}}
$$

where $\mu$ is a root of the following function:

$$
G(\mu): = \sum_{i=1}^n \left(\frac{\abs{A_{(i)}}\abs{B_{(i)}}}{\sqrt{\mu}}1_{\{0<\abs{A_{(i)}}\abs{B_{(i)}}<\mu \}} + 1_{\{\abs{A_{(i)}}\abs{B_{(i)}}\geq \mu \}} \right) -k
$$

\label{thm_sup:expected_frobenius}
\end{theorem_sup}
\begin{proof}
Note:
\begin{align*}
\E{\norm{A^\top B-\tilde{A}^\top \tilde{B}}_F^2} &= \sum_{i=1}^m\sum_{j=1}^p\E{\left(A^\top B-\tilde{A}^\top \tilde{B}\right)_{i,j}^2} \\
&= \sum_{i=1}^m\sum_{j=1}^p\Var{\left(\tilde{A}^\top \tilde{B}\right)_{i,j}}
\end{align*}
Therefore, using Proposition~\ref{proposition:variance} we get:
\begin{align*}
\E{\norm{A^\top B-\tilde{A}^\top \tilde{B}}_F^2} &=\sum_{i=1}^m\sum_{j=1}^p\sum_{t=1}^n\frac{1-p_t}{p_t}A_{t,i}^2B_{t,j}^2 \\
&=\sum_{t=1}^n\frac{1-p_t}{p_t}\left(\sum_{i=1}^mA_{t,i}^2\right)\left(\sum_{j=1}^pB_{t,j}^2\right) \\
&= \sum_{t=1}^n\frac{1-p_t}{p_t}\abs{A_{(t)}}^2\abs{B_{(t)}}^2
\end{align*}
Let us now find the optimal sampling probabilities that minimize the Frobenius error. Define the function:
$$
f(p_1,p_2,...,p_n)=\sum_{t=1}^n\frac{1-p_t}{p_t}\abs{A_{(t)}}^2\abs{B_{(t)}}^2
$$
We can now consider the optimization problem:
\begin{align*}
\min_{p_1,...,p_n} \quad & f(p_1,...,p_n) \\
\textrm{s.t} \quad & p_i-1 \leq 0 \\
    & -p_i \leq 0  \\
    & \sum_{i=1}^np_i - k = 0 
\end{align*}

We define the Lagrangian as:
\begin{align*}
& L(p_1,...,p_n,\lambda_1,...,\lambda_n,\nu_1,...,\nu_n,\mu)\triangleq \\ 
& f(p_1,p_2,...,p_k)+\sum_{i=1}^n\lambda_i\left(p_i-1\right)-\sum_{i=1}^n\nu_ip_i+\mu\left(\sum_{i=1}^np_i-k\right)
\end{align*}
where $\lambda_i\geq 0$, $\nu_i\geq 0$ and $\mu\in\mathbb{R}$.

Applying KKT stationarity condition:
$$
0=\frac{\partial}{\partial p_i}L=-\frac{1}{p_i^2}\abs{A_{(i)}}^2\abs{B_{i)}}^2+\lambda_i-\nu_i+\mu=0
$$ 
Therefore:
$$
p_i=\frac{\abs{A_{(i)}}\abs{B_{(i)}}}{\sqrt{\lambda_i-\nu_i+\mu}}
$$

Next we divide into 3 cases,\\
\textbf{Case 1: If $p_i\in(0,1)$:} In this case due to complementary-slackness we obtain $\lambda_i = \nu_i=0$, and therefore,
$$
p_i = \frac{\abs{A_{(i)}}\abs{B_{(i)}}}{\sqrt{\mu}}
$$
\\
\textbf{Case 2: If $p_i = 1$:} In this case due to complementary-slackness we obtain $\nu_i=0$, and therefore,
$$
1 = p_i = \frac{\abs{A_{(i)}}\abs{B_{(i)}}}{\sqrt{\mu+\lambda_i}}
$$
\\
\textbf{Case 3: If $p_i = 0$:} In this case due to complementary-slackness we obtain $\lambda_i=0$, which implies that,
$$
0 = p_i = \frac{\abs{A_{(i)}}\abs{B_{(i)}}}{\sqrt{\mu-\nu_i}}
$$
but this can only happen if $\abs{A_{(i)}}\abs{B_{(i)}}=0$.

Combining the above we conclude that given $\mu$ one can write the solution as follows,

$$
p_i = \frac{\abs{A_{(i)}}\abs{B_{(i)}}}{\sqrt{\mu}}1_{\{0<\abs{A_{(i)}}\abs{B_{(i)}}<\mu \}} + 1_{\{\abs{A_{(i)}}\abs{B_{(i)}}\geq \mu \}}
$$

Now, in order to satisfy the equality conditions $\mu$ should satisfy the following equality,
$$
\sum_{i=1}^n \left(\frac{\abs{A_{(i)}}\abs{B_{(i)}}}{\sqrt{\mu}}1_{\{0<\abs{A_{(i)}}\abs{B_{(i)}}<\mu \}} + 1_{\{\abs{A_{(i)}}\abs{B_{(i)}}\geq \mu \}} \right) =k
$$
Now, one can actually find $\mu$ using bisection, To see this consider the following function,
$$
G(\mu): = \sum_{i=1}^n \left(\frac{\abs{A_{(i)}}\abs{B_{(i)}}}{\sqrt{\mu}}1_{\{0<\abs{A_{(i)}}\abs{B_{(i)}}<\mu \}} + 1_{\{\abs{A_{(i)}}\abs{B_{(i)}}\geq \mu \}} \right) -k
$$
And note that $G(\mu)$ is a one dimensional monotonically decreasing (actually non-increasing) function of $\mu$.

Also, if we sorts the  $\abs{A_{(i)}}\abs{B_{(i)}}$'s, i.e. $\abs{A_{(1)}}\abs{B_{(1)}}\leq \abs{A_{(2)}}\abs{B_{(2)}}\leq\ldots \abs{A_{(n)}}\abs{B_{(n)}}$, then given $j$ such that  $\mu\in(\abs{A_{(j)}}\abs{B_{(j)}},\abs{A_{(j+1)}}\abs{B_{(j+1)}})$, then we can find the exact value of $\mu$ from the equality constraints equation:
$$
\sum_{i=1}^n \left(\frac{\abs{A_{(i)}}\abs{B_{(i)}}}{\sqrt{\mu}}1_{\{0<\abs{A_{(i)}}\abs{B_{(i)}}<\mu \}} + 1_{\{\abs{A_{(i)}}\abs{B_{(i)}}\geq \mu \}} \right) =k
$$
\end{proof}

\begin{*corollary}
The sampling probabilities
\begin{align*}
p_i=\min\left\{\frac{k\abs{A_{(i)}}\abs{B_{(i)}}}{\sum_{j=1}^n\abs{A_{(j)}}\abs{B_{(j)}}},1\right\}
\end{align*}
are optimal if $k\leq \frac{\sum_{i=1}^n\abs{A_{(i)}}\abs{B_{(i)}}}{\max_i\abs{A_{(i)}}\abs{B_{(i)}}}$
\end{*corollary}

\begin{proof}
As a simpler, sub-optimal solution for the above optimization problem we propose the following relaxation. First, we solve the optimization problem without the inequality conditions:
\begin{align*}
0\leq p_i\leq 1
\end{align*}
Then, for each optimal $p_i^*$ we clamp the value between the range $[0,1]$. This allows us to comply with the inequality conditions that allows to treat $p_i$ as a parameter to Bernoulli distribution at the expense of relaxing the constraint on the sum of the parameters $p_i$, leading to potentially sub-optimal solution.

As the first step, we therefore solve the problem:
\begin{align*}
\min_{p_1,...,p_n} \quad & f(p_1,...,p_n) \\
\textrm{s.t} \quad & \sum_{i=1}^np_i - k
\end{align*}

To minimize $f$ subject to the constraint $\sum_{i=1}^n p_i = k$ we use the Lagrange multiplier $\lambda$ and define the function:
$$
g(p_1,p_2,...,p_n)=f(p_1,p_2,...,p_n)+\lambda\left(\sum_{i=1}^n p_i-k \right)
$$
Deriving and equaling to zero we get:
$$
0 = \frac{\partial g}{\partial p_i}=-\frac{1}{p_i^2}\abs{A_{(i)}}^2\abs{B_{i)}}^2+\lambda
$$
Therefore:
$$
p_i=\frac{\abs{A_{(i)}}\abs{B_{(i)}}}{\sqrt{\lambda}}
$$
Substituting in $\sum_{i=1}^n p_i = k$:
$$
\sum_{i=1}^n \frac{\abs{A_{(i)}}\abs{B_{i)}}}{\sqrt{\lambda}} = k
$$
$$
\sqrt{\lambda} = \frac{\sum_{i=1}^n\abs{A_{(i)}}\abs{B_{(i)}}}{k}
$$
And therefore we get:
$$
p_i=\frac{k\abs{A_{(i)}}\abs{B_{(i)}}}{\sum_{i=1}^n\abs{A_{(i)}}\abs{B_{(i)}}}
$$
And the final result after clamping would be:
$$
p_i=\min\left\{\frac{k\abs{A_{(i)}}\abs{B_{(i)}}}{\sum_{i=1}^n\abs{A_{(i)}}\abs{B_{(i)}}},1\right\}
$$

Note that this solution yields $p_i\geq 0$, satisfying one of the original inequality conditions. What about $p_i\leq 1$?

If 
$$
k\leq \frac{\sum_{i=1}^n\abs{A_{(i)}}\abs{B_{(i)}}}{\max_i\abs{A_{(i)}}\abs{B_{(i)}}}
$$
then the second inequality conditions holds as well and the solution is indeed the optimal solution to the original problem.

Substituting in the expression for the Frobenius error we get:
\begin{align*}
& \E{\norm{A^\top B-\tilde{A}^\top \tilde{B}}_F^2} =\sum_{t=1}^n\frac{1-p_t}{p_t}\abs{A_{(t)}}^2\abs{B_{(t)}}^2 \\
&=\frac{1}{k}\left(\sum_{i=1}^n\abs{A_{(i)}}\abs{B_{(i)}}\right)^2-\sum_{i=1}^n\abs{A_{(i)}}^2\abs{B_{(i)}}^2
\end{align*}
\end{proof}

The following theorem yields high probability bounds for the Frobenius and spectral norms for the Bernoulli-CRS algorithm:
\begin{theorem_sup}
Let $A\in\mathbb{R}^{n\times m}$ and $B\in\mathbb{R}^{n\times p}$. Let $\tilde{A},\tilde{B}$ be the sampled matrices according to the Bernoulli-CRS algorithm described above.
Denote
$$
R \triangleq \max_i\norm{A^{\top(i)} B_{(i)}}
$$
and
$$
\sigma^2 \triangleq \frac{1}{k}\left(\sum_{i=1}^n e_i\abs{A_{(i)}}\abs{B_{(i)}}\right)^2-\sum_{i=1}^n e_i\abs{A_{(i)}}^2\abs{B_{(i)}}^2
$$
then, for all $t\geq 0$:
$$
\mathbb{P}\left\{\norm{A^\top B - \tilde{A}^\top \tilde{B}}\geq t\right\} \leq (m+p)\cdot \exp\left(\frac{-t^2/2}{\sigma^2+Rt/3}\right)
$$
$$
\mathbb{P}\left\{\norm{A^\top B - \tilde{A}^\top \tilde{B}}_F\geq t\right\} \leq (m+p)^{3/2}\cdot \exp\left(\frac{-t^2/2}{\sigma^2+Rt/3}\right)
$$
\label{thm_sup:bernoulli_berenstein}
\end{theorem_sup}
\begin{proof}
The Matrix Bernstein concentration inequality states:

\begin{*theorem}[\textbf{Matrix Bernstein \cite{tropp2012user}}]
Consider a finite sequence $\{\mathbf{Z}_k\}$ of independent, random matrices with dimensions $d_1$ $\times$ $d_2$. Assume that each random matrix satisfies
$$
\E{\mathbf{Z}_k}=0 \quad \text{and} \quad \norm{\mathbf{Z}_k}\leq R \quad \text{almost surely}.
$$
Define
$$
\sigma^2 \triangleq \max\left\{\norm{\sum_k\E{\mathbf{Z}_k\mathbf{Z}_k^\top }},\norm{\sum_k\E{\mathbf{Z}_k^\top \mathbf{Z}_k}} \right\}
$$
Then, for all $t\geq 0$,
$$
\mathbb{P}\left\{\norm{\sum_k\mathbf{Z}_k}\geq t\right\} \leq (d_1+d_2)\cdot \exp\left(\frac{-t^2/2}{\sigma^2+Rt/3}\right)
$$.
\end{*theorem}

In our sampling algorithm, we can define:
$$
\mathbf{Z}_k \triangleq A_{(k)}^\top B_{(k)}-\frac{1}{p_k}K_{k,k}A_{(k)}^\top B_{(k)}
$$
when $K_{k,k}$ is a Bernoulli random variable with parameter $p_k$ as defined above. It is clear that $\E{\mathbf{Z}_k}=0$. 

Also, let us define:
$$
R \triangleq \max_k\norm{A_{(k)}^\top B_{(k)}}
$$
so it it also clear that $\norm{\mathbf{Z}_k}\leq R$. 

By construction, $\{\mathbf{Z}_k\}$ are independent.

We can also define:
\begin{align*}
\sigma^2 &\triangleq \max\left\{\norm{\sum_k\E{\mathbf{Z}_k\mathbf{Z}_k^\top }},\norm{\sum_k\E{\mathbf{Z}_k^\top \mathbf{Z}_k}} \right\} \\
&=\max\left\{\norm{\E{\sum_k\mathbf{Z}_k\mathbf{Z}_k^\top }},\norm{\E{\sum_k\mathbf{Z}_k^\top \mathbf{Z}_k}} \right\} \\
&= \max\bigg \{\norm{\E{(A^\top B-\tilde{A}^\top \tilde{B})(A^\top B-\tilde{A}^\top \tilde{B})^\top }}, \\
&\norm{\E{(A^\top B-\tilde{A}^\top \tilde{B})^\top (A^\top B-\tilde{A}^\top \tilde{B})}} \bigg \} \\
& \leq \max\bigg \{\Tr\left(\E{(A^\top B-\tilde{A}^\top \tilde{B})(A^\top B-\tilde{A}^\top \tilde{B})^\top }\right), \\
& \Tr\left(\E{(A^\top B-\tilde{A}^\top \tilde{B})^\top (A^\top B-\tilde{A}^\top \tilde{B})}\right)\bigg \} \\
& = \E{\norm{A^\top B-\tilde{A}^\top \tilde{B}}_F^2} \\
&= \frac{1}{k}\left(\sum_{i=1}^n e_i\abs{A_{(i)}}\abs{B_{(i)}}\right)^2-\sum_{i=1}^n e_i\abs{A_{(i)}}^2\abs{B_{(i)}}^2
\end{align*}
where we used the linearity of expectation and trace, the property $\norm{A}\leq\Tr[A]$ for positive semi-definite matrices and the expected Frobenius norm from Theorem~\ref{thm_sup:expected_frobenius}.

The bound on the spectral norm follows immediately from the Matrix Berenstein inequality.

Using the property:
$$
\norm{A}_F\leq\sqrt{r}\norm{A}
$$ 
we get the similar result for the Frobenius norm, factored by $\sqrt{m+p}$.
\end{proof}

\subsection{Proofs for Section~\ref{section:conv_details_sup} - Approximating Convolutions}
\label{section:conv_proofs}
The following proofs go along the same lines of \citep{drineas2006fast}, generalizing them to multi-channel convolutions (zero-padding assumed).
\begin{lemma_sup}
\label{lemma_conv_estimator}
Suppose $I\in\mathbb{R}^{IW\times IC}_{B\times IH}, K\in\mathbb{R}^{IC\times OC}_{KW\times KW}, 1\leq k\leq IC, \{p_i\}_{i=1}^{IC}$ is a probability distribution over $\{1,...,IC\}$ and $\{i_t\}_{t=1}^{k}$ are such that $i_t\in\{1,...,IC\}$.

Let $O\in\mathbb{R}^{OW\times OC}_{B\times OH}=I*K$ be the multi-channel convolution of $I,K$ as defined in~\eqref{eq_convolution} and let  $\tilde{O}$ be its approximation by sampling $k$ input channels as defined in~\eqref{eq_approximate_convolution_tensor}. Then:
\begin{align*}
\E{\tilde{O}}=O
\end{align*}
\end{lemma_sup}
\begin{proof}
We show that every $b,oh,ow,oc$ satisfies $\E{\tilde{O}_{b,oh}^{ow,oc}}=O_{b,oh}^{ow,oc}$. 

For $t\in\{1,...,k\}$, define $X_{t}=(\frac{I^{[i_{t}]}*K_{[i_{t}]}}{p_{i_{t}}})_{b,oh}^{ow,oc}$. 

Using~\eqref{eq_approximate_convolution_tensor} we can write $\tilde{O}_{b,oh,ow,oc}=\sum_{t=1}^{k}\frac{1}{k}X_{t}$. 

Taking the expectation, we get:
\begin{align}
(\E{\tilde{O}})_{b,oh}^{ow,oc}=\E{\sum_{t=1}^{k}\frac{1}{k}X_{t}} =\E{X_{t}}=\sum_{i=1}^{IC}p_{i}\cdot\frac{(I^{[i]}*K_{[i]})_{b,oh}^{ow,oc}}{p_{i}}=O_{b,oh}^{ow,oc}
\label{eq_conv_unbiased}
\end{align}
\end{proof}

\begin{lemma_sup}
Suppose the same as Lemma~\ref{lemma_conv_estimator}. Then:
\begin{equation*}
\begin{split}
\Var{\tilde{O}_{b,oh}^{ow,oc}}
&=\frac{1}{k}\sum_{i=1}^{IC}\frac{1}{p_i}\sum_{h=1}^{KH}\sum_{w=1}^{KW}(I_{b,oh+h-1}^{ow+w-1,i})^{2}(K_{h,w}^{i,oc})^{2} \\
&+ \frac{1}{k}\sum_{i=1}^{IC}\frac{1}{p_i}\sum_{\substack{h,h'=1\\h\neq h'}}^{KH}\sum_{\substack{w,w'=1\\w\neq w'}}^{KW}I_{b,oh+h-1}^{ow+w-1,i}I_{b,oh+h'-1}^{ow+w'-1,i}K_{h,w}^{i,oc}K_{h',w'}^{i,oc} \\
&-\frac{1}{k}(O_{b,oh}^{ow,oc})^{2}
\end{split}
\end{equation*}
\label{lemma_conv_var}
\end{lemma_sup}
\begin{proof}
Define $X_t$ as in Lemma \ref{lemma_conv_estimator}. From~\eqref{eq_approximate_convolution_tensor} and the independence of different $X_t$:
\begin{align}
\Var{\tilde{O}_{b,oh}^{ow,oc}}=\Var{\sum_{t=1}^{k}\frac{1}{k}X_{t}}=\frac{1}{k}\Var{X_t}=\frac{1}{k}(\E{X_t^2}-\E{X_t}^2)
\label{eq_conv_var}
\end{align}

\begin{align}
\begin{split}
\E{X_t^2}&=\sum_{i=1}^{IC}p_{i}\cdot\frac{((I^{[i]}*K_{[i]})_{b,oh}^{ow,oc})^2}{p_{i}^2} \\
&=\sum_{i=1}^{IC}\frac{1}{p_i}\left(\sum_{h=1}^{KH}\sum_{w=1}^{KW}I_{b,oh+h-1}^{ow+w-1,i}K_{h,w}^{i,oc}\right)^{2}
\end{split}
\label{eq_conv_2nd_moment}
\end{align}
From~\eqref{eq_conv_unbiased} we get $\E{X_t}=O$.

Substituting both expressions in~\eqref{eq_conv_var} and expanding concludes the proof.
\end{proof}

\begin{lemma_sup}
Suppose the same as Lemma~\ref{lemma_conv_estimator}. Then:
\begin{align*}
\E{\big\Vert O-\tilde{O}\big\Vert_{F}^2}=\sum_{i=1}^{IC}\frac{\big\Vert I^{[i]}\big\Vert_{F}^2\cdot\big\Vert K_{[i]}\big\Vert_{F}^2-E_{IK}^i+R_{IK}^i}{kp_i}
-\frac{1}{k}\big\Vert O\big\Vert_{F}^2
\end{align*}
where 
\begin{align*}
\begin{split}
E_{IK}^i&=\sum_{b=1}^{B}\sum_{\substack{oh,ow\text{ s.t} \\oh<KH \text{ or} \\ ow<KW}}\sum_{oc=1}^{OC}\sum_{h=1}^{KH}\sum_{\substack{h,w\text{ s.t} \\h>oh \text{ or} \\ w>ow}}^{KH,KW} (I_{b,oh}^{ow,i})^2(K_{h,w}^{i,oc})^2
\\
R_{IK}^i&=\sum_{b=1}^B\sum_{oh=1}^{OH}\sum_{ow=1}^{OW}\sum_{oc=1}^{OC}\sum_{\substack{h,h'=1\\h\neq h'}}^{KH}\sum_{\substack{w,w'=1\\w\neq w'}}^{KW}I_{b,oh+h-1}^{ow+w-1,i}I_{b,oh+h'-1}^{ow+w'-1,i}K_{h,w}^{i,oc}K_{h',w'}^{i,oc}
\end{split}
\end{align*}
The expected error is minimized when the sampling probabilities are:
\begin{align*}
p_i=\frac{\sqrt{\big\Vert I^{[i]}\big\Vert_{F}^2\cdot\big\Vert 
K_{[i]}\big\Vert_{F}^2-E_{IK}^i+R_{IK}^i}}{\sum_{j=1}^{IC}
\sqrt{\big\Vert I^{[j]}\big\Vert_{F}^2\cdot\big\Vert K_{[j]}\big
\Vert_{F}^2-E_{IK}^j+R_{IK}^j}}
\end{align*}
\begin{remark}
We use here the Frobenius norm in its generalization for tensors. For a tensor T of rank r:
\begin{align*}
\big\Vert T\big\Vert _{F}=\sqrt{\sum_{j_{1},j_{2},...,j_{r}}T_{j_{1},j_{2},...,j_{r}}^{2}}
\end{align*}
\end{remark}
\label{lemma_conv_frobenius}
\end{lemma_sup}
\begin{proof}
Note that:
\begin{align*}
\E{\big\Vert O-\tilde{O}\big\Vert_{F}^2}=\sum_{b=1}^B\sum_{oh=1}^{OH}\sum_{ow=1}^{OW}\sum_{oc=1}^{OC}\E{((O-\tilde{O})_{b,oh}^{ow,oc})^2}=\sum_{b=1}^B\sum_{oh=1}^{OH}\sum_{ow=1}^{OW}\sum_{oc=1}^{OC}\Var{\tilde{O}_{b,oh}^{ow,oc}}
\end{align*}
Substituting the result from Lemma~\ref{lemma_conv_var}:
\begin{align}
\begin{split}
&\E{\big\Vert O-\tilde{O}\big\Vert_{F}^2}
=\sum_{i=1}^{IC}\frac{1}{kp_i}\sum_{b=1}^B\sum_{oh=1}^{OH}\sum_{ow=1}^{OW}\sum_{oc=1}^{OC}\sum_{h=1}^{KH}\sum_{w=1}^{KW}(I_{b,oh+h-1}^{ow+w-1,i})^2(K_{h,w}^{i,oc})^2 \\
&+ \sum_{i=1}^{IC}\frac{1}{kp_i}\sum_{b=1}^B\sum_{oh=1}^{OH}\sum_{ow=1}^{OW}\sum_{oc=1}^{OC}\sum_{\substack{h,h'=1\\h\neq h'}}^{KH}\sum_{\substack{w,w'=1\\w\neq w'}}^{KW}I_{b,oh+h-1}^{ow+w-1,i}I_{b,oh+h'-1}^{ow+w'-1,i}K_{h,w}^{i,oc}K_{h',w'}^{i,oc} \\
&-\frac{1}{k}\sum_{b=1}^B\sum_{oh=1}^{OH}\sum_{ow=1}^{OW}\sum_{oc=1}^{OC}(O_{b,oh}^{ow,oc})^{2}
\end{split}
\label{eq_nasty_indices}
\end{align}
This expression includes 3 terms. The first involves products between each element of $I^{[i]}$ and all the corresponding entries in $K_{[i]}$, except for the upper and left edges of $I^{[i]}$. We therefore add and subtract the correction term ${E_{IK}^{i}}$ to get:
\begin{align*}
\begin{split}
&\sum_{i=1}^{IC}\frac{1}{kp_i}\sum_{b=1}^B\sum_{oh=1}^{OH}\sum_{ow=1}^{OW}\sum_{oc=1}^{OC}\sum_{h=1}^{KH}\sum_{w=1}^{KW}(I_{b,oh+h-1}^{ow+w-1,i})^2(K_{h,w}^{i,oc})^2 \\
&=\sum_{i=1}^{IC}\frac{1}{kp_i}\left(\left(\sum_{b=1}^B\sum_{oh=1}^{OH}\sum_{ow=1}^{OW}(I_{b,oh+h-1}^{ow+w-1,i})^2\right)\left(\sum_{oc=1}^{OC}\sum_{h=1}^{KH}\sum_{w=1}^{KW}(K_{h,w}^{i,oc})^2\right)-E_{IK}^i\right) \\
&= \sum_{i=1}^{IC}\frac{\big\Vert I^{[i]}\big\Vert_{F}^2\cdot\big\Vert K_{[i]}\big\Vert_{F}^2-E_{IK}^i}{kp_i}
\end{split}
\end{align*}  
The second term is \begin{math}\sum_{i=1}^{IC}\frac{1}{kp_i}R_{IK}^{i}\end{math}.

The third term can be written as \begin{math}\frac{1}{k}\sum_{b=1}^B\sum_{oh=1}^{OH}\sum_{ow=1}^{OW}\sum_{oc=1}^{OC}(O_{b,oh,ow,oc})^{2}=\frac{1}{k}\big\Vert O\big\Vert_F^2\end{math}

Substituting these terms in~\eqref{eq_nasty_indices} yields the result of~\eqref{lemma_conv_frobenius}.

To find \begin{math}\{p_i\}_{i=1}^{IC}\end{math} that minimize the expression in~\eqref{lemma_conv_frobenius} it is enough to minimize the function \begin{math}f=\sum_{i=1}^{IC}\frac{\alpha_i^2}{p_i}\end{math} under the constraints \begin{math}\sum p_i=1\end{math} and \begin{math}p_i>0\end{math}. We can write the numerator as \begin{math}\alpha_i^2\end{math} because the expression in~\eqref{eq_conv_2nd_moment} is non-negative.

This minimization problem has a straightforward solution in Lemma 4 of \citep{drineas2006fast}, which is \begin{math}p_i=\frac{\alpha_i}{\sum_{j=1}^{IC}\alpha_j}\end{math}.

In our case, \begin{math}\alpha_i=\sqrt{\big\Vert I^{[i]}\big\Vert_{F}^2\cdot\big\Vert K_{[i]}\big\Vert_{F}^2-E_{IK}^i+R_{IK}^i}\end{math}, and therefore the optimal probabilities are:
\begin{equation*}
p_i=\frac{\sqrt{\big\Vert I^{[i]}\big\Vert_{F}^2\cdot\big\Vert 
K_{[i]}\big\Vert_{F}^2-E_{IK}^i+R_{IK}^i}}{\sum_{j=1}^{IC}
\sqrt{\big\Vert I^{[j]}\big\Vert_{F}^2\cdot\big\Vert K_{[j]}\big
\Vert_{F}^2-E_{IK}^j+R_{IK}^j}}
\end{equation*}

The terms \begin{math}E_{IK}^i,R_{IK}^i\end{math} emerge for convolutions when the kernel spatial dimensions are greater than one. However, computing them is too expensive, precluding efficient implementation of the approximate version. We therefore omit them and verify empirically whether the resulting norm-proportional probabilities: 
\begin{align*}
p_i=\frac{\big\Vert I^{[i]}\big\Vert_{F}\cdot\big\Vert 
K_{[i]}\big\Vert_{F}}{\sum_{j=1}^{IC}
\big\Vert I^{[j]}\big\Vert_{F}\cdot\big\Vert K_{[j]}\big
\Vert_{F}}
\end{align*}

yield better results than the uniform sampling. Intuitively, in some (common) cases these terms are much smaller than $\big\Vert I^{[i]}\big\Vert_{F}^2\cdot\big\Vert K_{[i]}\big\Vert_{F}^2$, so their omission does not significantly impact the final outcome. $E_{IK}^i$ amounts to the outer spatial dimensions of the input not being convolved with the entire kernel, so it is likely to be smaller than the Frobenius norm of the whole input. $R_{IK}^i$ is the sum of products of different input and	kernel entries. If different kernels are lowly-correlated and weights are centered around zero, the sum will include terms of similar magnitudes but opposite signs.

\end{proof}

\section{Implementation Details}
\label{section:implementation_details}
All single-node results were obtained using 2.2GHz Intel Xeon Silver 4210 CPU with four NVidia V100 GPUs with 32GB of memory. Wall-time speedup were measured when running with a single GPU, except ResNet-152 where 2 GPUs are used due to memory capacity. We used PyTorch version 1.7.0 with CUDA 10.1 and Python version 3.6.9.

\subsection{MLP for MNIST}
The MNIST dataset \citep{lecun1998gradient} includes 60K training examples and 10K test examples. We use 5K as validation set. Each example is a $28\times 28$ gray-scale image of a handwritten digit.

Our MLP model contains the following layers:
\begin{itemize}
\item $784\times 500$ fully-connected layer with RELU activations.
\item $500\times 500$ fully-connected layer with RELU activations.
\item $500\times 10$ fully-connected layer with RELU activations.
\item Log Softmax
\end{itemize}

We use the Adam optimizer \citep{kingma2014adam} with default parameters (learning rate=0.001,$\beta_{1}=0.9$,$\beta_{2}=0.999$,$\epsilon=1e-08 $). As loss function we use negative log likelihood. We use minibatch size of 50 and train the model for 20 epochs.

We apply sampling to all the fully connected layers. When sampling in the backward pass, we do not reduce the batch dimension below 10 in the weight gradient computation.

\begin{figure}[h]
  \subfigure[Sampling in forward pass]{\includegraphics[width=0.5\linewidth]{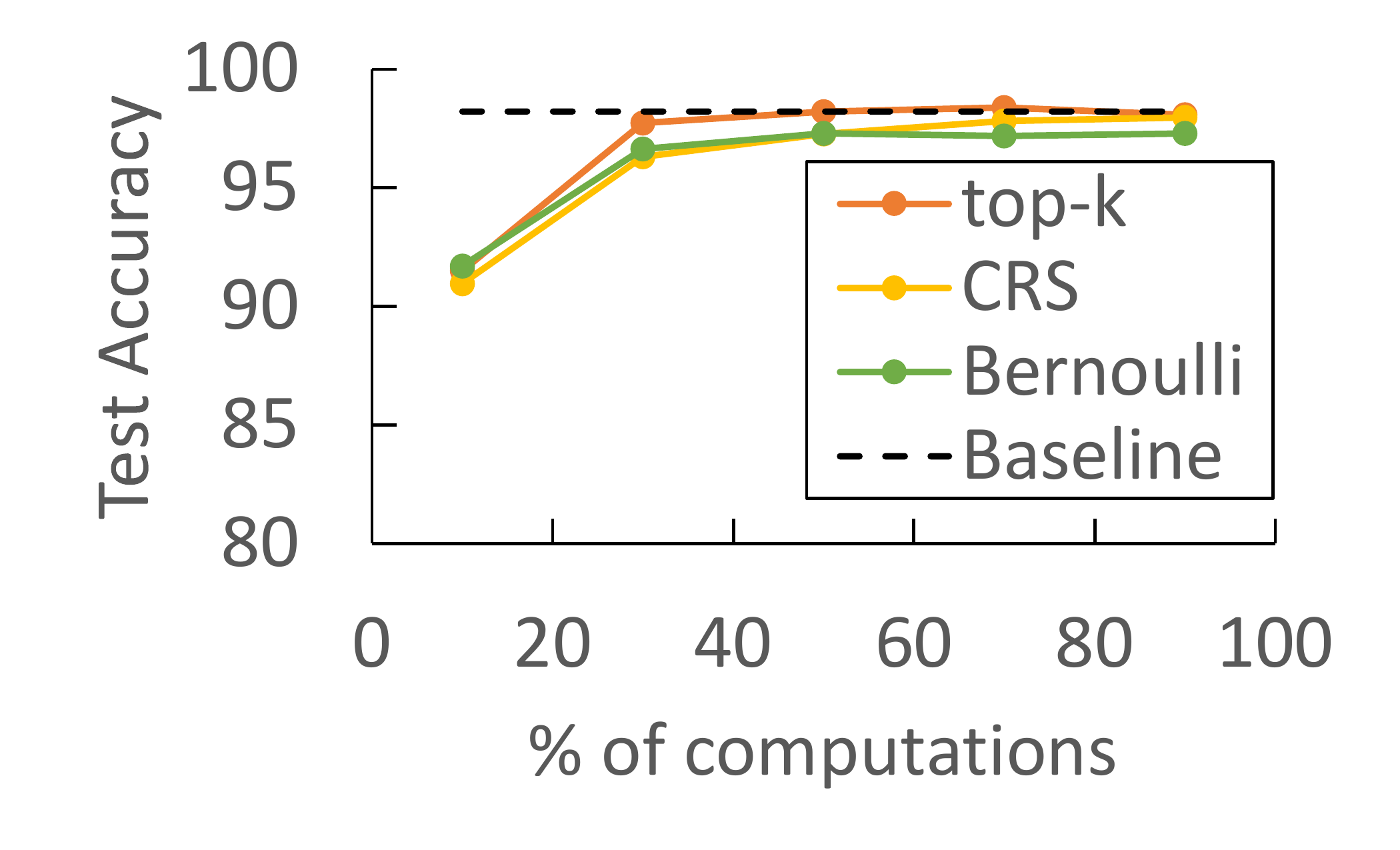}
  \label{fig_sup:mnist_mlp_forward}}~
  \subfigure[Sampling in backward pass]{\includegraphics[width=0.5\linewidth]{fig_mnist_mlp_backward.pdf}
  \label{fig_sup:mnist_mlp_backward}}
  \caption{MNIST test accuracy for MLP, under different approximating algorithms and different sampling ratios}
  \label{fig_mlp}
\end{figure}

Figure~\ref{fig_sup:mnist_mlp_forward} shows the MNIST test accuracy for different sampling algorithms and sampling ratios in the forward pass. We observe that top-$k$ performs the best. Figure~\ref{fig_sup:mnist_mlp_backward} shows the same when approximations are applied in the backward pass only. In this case, all sampling algorithms are similar when performing above 30\% of the backward pass computations.

\subsection{CNN for MNIST}
The network is composed of the following layers:
\begin{itemize}
\item $5\times 5\times 32$ convolution layer with RELU activation, followed by $2\times 2$ max pooling.
\item $5\times 5\times 64$ convolution layer with RELU activation, followed by $2\times 2$ max pooling.
\item Dropout layer with $p=0.5$.
\item $3136\times 1024$ fully connected layer with RELU activation.
\item $1024\times 10$ fully connected layer.
\item Dropout layer with $p=0.5$.
\item Log Softmax
\end{itemize}

The model is trained using Adam optimizer with default parameters (learning rate=0.001,\begin{math}\beta_{1}=0.9\end{math},\begin{math}\beta_{2}=0.999\end{math},\begin{math}\epsilon=1e-08 \end{math}) and negative log likelihood  loss. We use minibatch size of 50 and train the model for 20 epochs.

We apply sampling to the convolutional layers. When sampling in the backward pass, we do not reduce the batch dimension below 10 in the weight gradient computation.

\begin{figure}[h]
  \subfigure[Sampling in forward pass]{\includegraphics[width=0.5\linewidth]{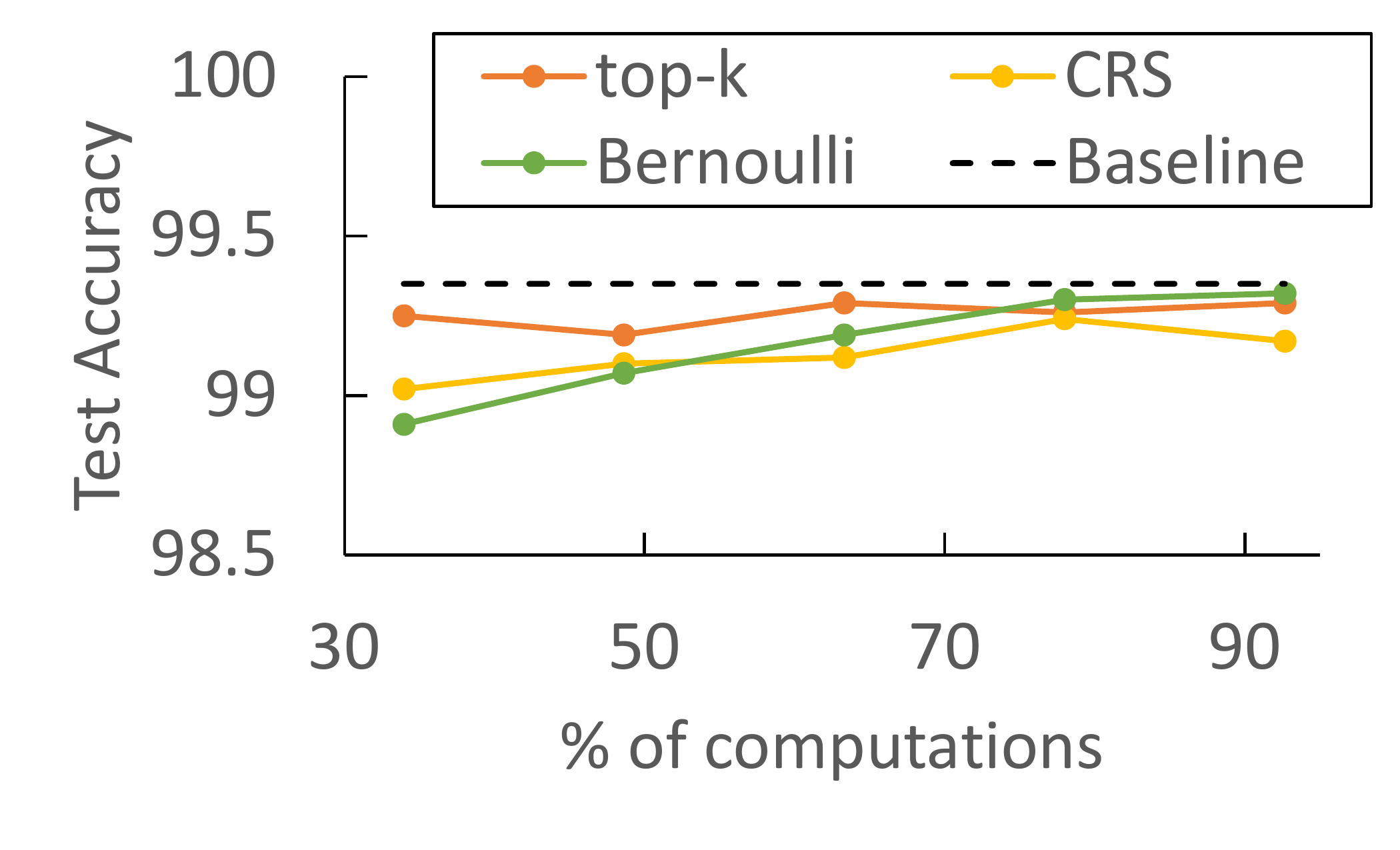}
  \label{fig_sup:mnist_cnn_forward}}~
  \subfigure[Sampling in backward pass]{\includegraphics[width=0.5\linewidth]{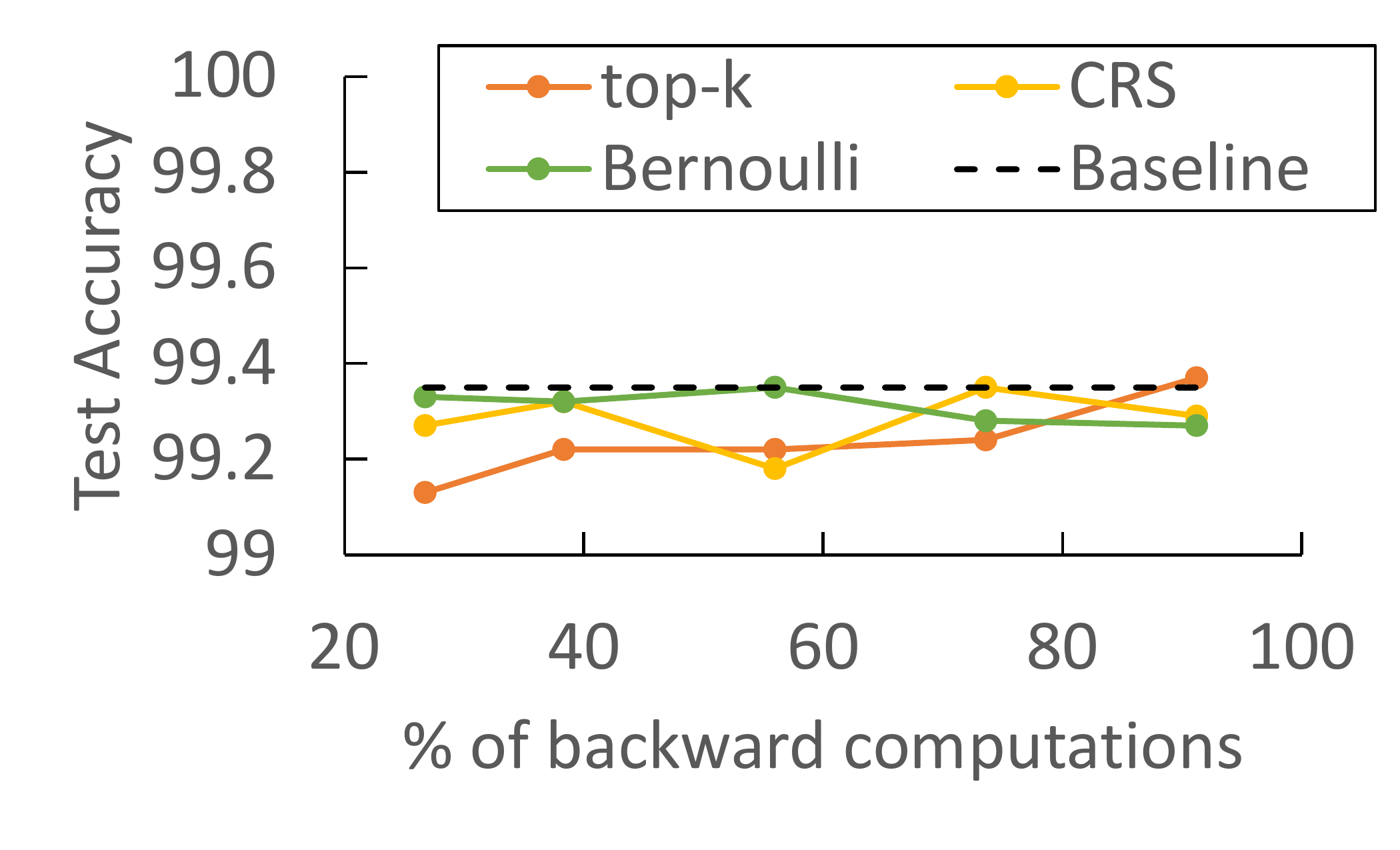}
  \label{fig_sup:mnist_cnn_backward}}
  \caption{MNIST test accuracy for CNN, under different approximating algorithms and different sampling ratios}
  \label{fig_cnn}
\end{figure}

Figure~\ref{fig_sup:mnist_cnn_forward} shows the MNIST test accuracy for different sampling algorithms and sampling ratios in the forward pass. We observe that top-$k$ performs the best. Figure~\ref{fig_sup:mnist_cnn_backward} shows the same when approximations are applied in the backward pass only. In this case, all sampling algorithms are similar when performing above 30\% of the backward pass computations.

\subsection{Wide ResNet-28-10 for CIFAR-10}
The CIFAR-10 dataset \citep{krizhevsky2009learning} consists of \begin{math}32\times 32\end{math} color images from 10 classes, split into 50K training set and 10K test set. 

For WRN-28-10 \citep{zagoruyko2016wide} we use the implementation in \url{https://github.com/meliketoy/wide-resnet.pytorch}, avialable under MIT License.

WRN-28-10 includes the following layers:
\begin{itemize}
\item conv1 - \begin{math}3\times 3\times 16\end{math} input convolution layer
\item conv2 - eight \begin{math}3\times 3\times 160\end{math} convolution layers
\item conv3 - eight \begin{math}3\times 3\times 320\end{math} convolution layers
\item conv4 - eight \begin{math}3\times 3\times 640\end{math}
convolution layers
\item Batch normalization, \begin{math}8\times 8\end{math} Average pooling, fully connected+softmax layers.
\end{itemize}

Every two subsequent convolution layers are followed by a residual connection that adds the input to these layers to the result. the first convolution conv3 and conv4 has a stride of 2, halving the spatial dimensions. For additional details see \citep{zagoruyko2016wide}.

Image preprocessing includes padding to 36x36 and random crop, horizontal flipping and per-image whitening. The optimizer is Momentum SGD with momentum=0.9 and 5e-4 weight decay. Learning rate is 0.15 for the first 60 epochs, 0.03 until epoch 120, 0.006 until epoch 160 and 0.0012 afterwards. We use batch size of 256, cross-entropy loss and train the model for 200 epochs.

We apply sampling to the convolutional layers except the first layer due to the small number of input channels (3) and the single fully-connected layer which amounts only to 0.01\% of the total computations in WRN-28-10. When sampling in the backward pass, we do not reduce the batch dimension below 10 in the weight gradient computation.

\begin{figure}[h]
  \subfigure[Sampling in forward pass]{\includegraphics[width=0.5\linewidth]{fig_wrn_cifar10_forward.pdf}
  \label{fig_sup:wrn_cifar10_forward}}~
  \subfigure[Sampling in backward pass]{\includegraphics[width=0.5\linewidth]{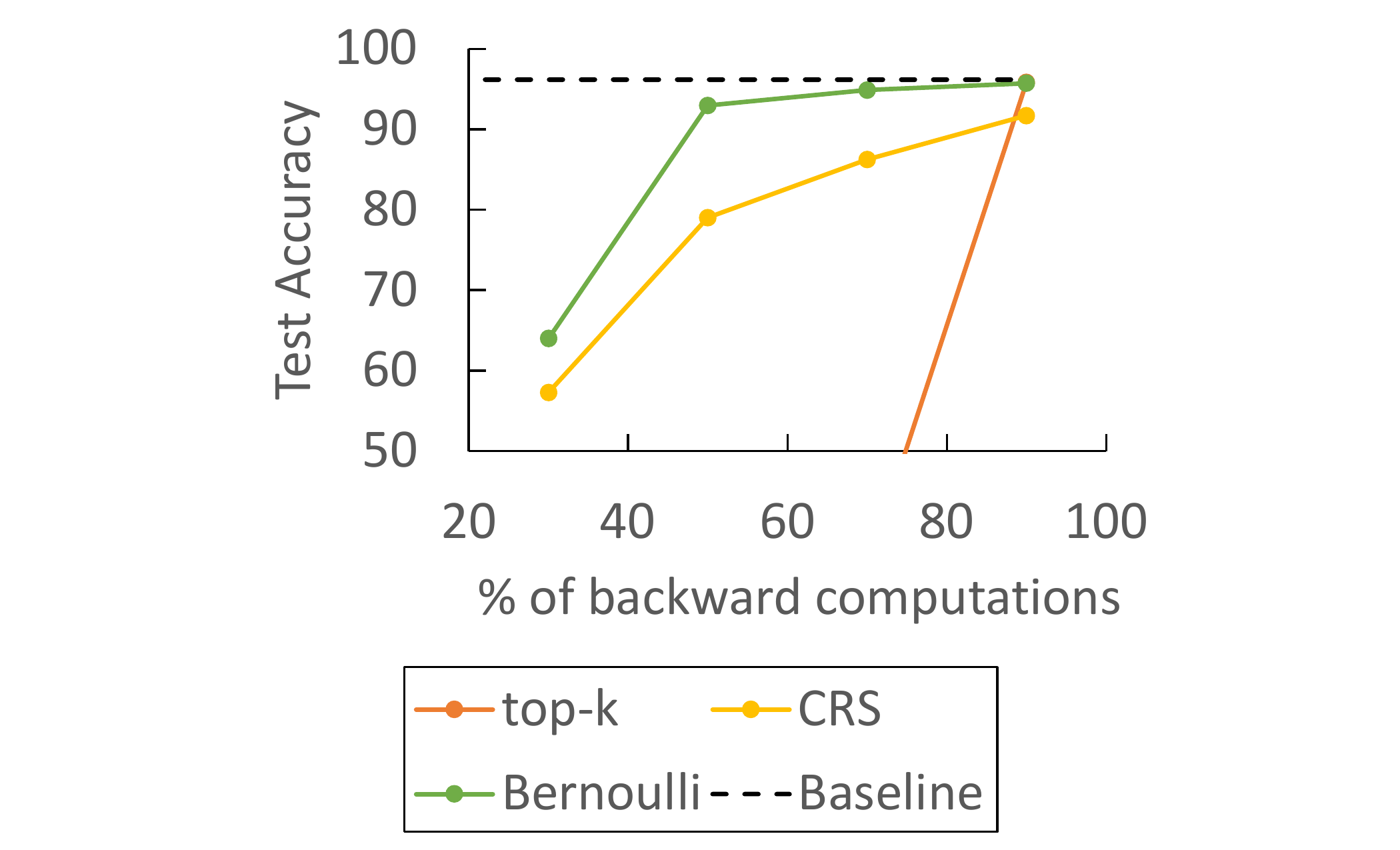}
  \label{fig_sup:wrn_cifar10_backward}}
  \caption{CIFAR-10 test accuracy for WRN-28-10, under different approximating algorithms and different sampling ratios}
  \label{fig_wrn}
\end{figure}

Figure~\ref{fig_sup:wrn_cifar10_forward} shows the CIFAR-10 test accuracy for different sampling algorithms and sampling ratios in the forward pass. We observe that top-$k$ performs the best. Figure~\ref{fig_sup:wrn_cifar10_backward} shows the same when approximations are applied in the backward pass only. In this case, Bernoulli-CRS performs the best but is still below 1\% of the baseline accuracy until 90\% sampling ratio. 

\begin{figure}[h]
  \centering
  \includegraphics[width=0.5\linewidth]{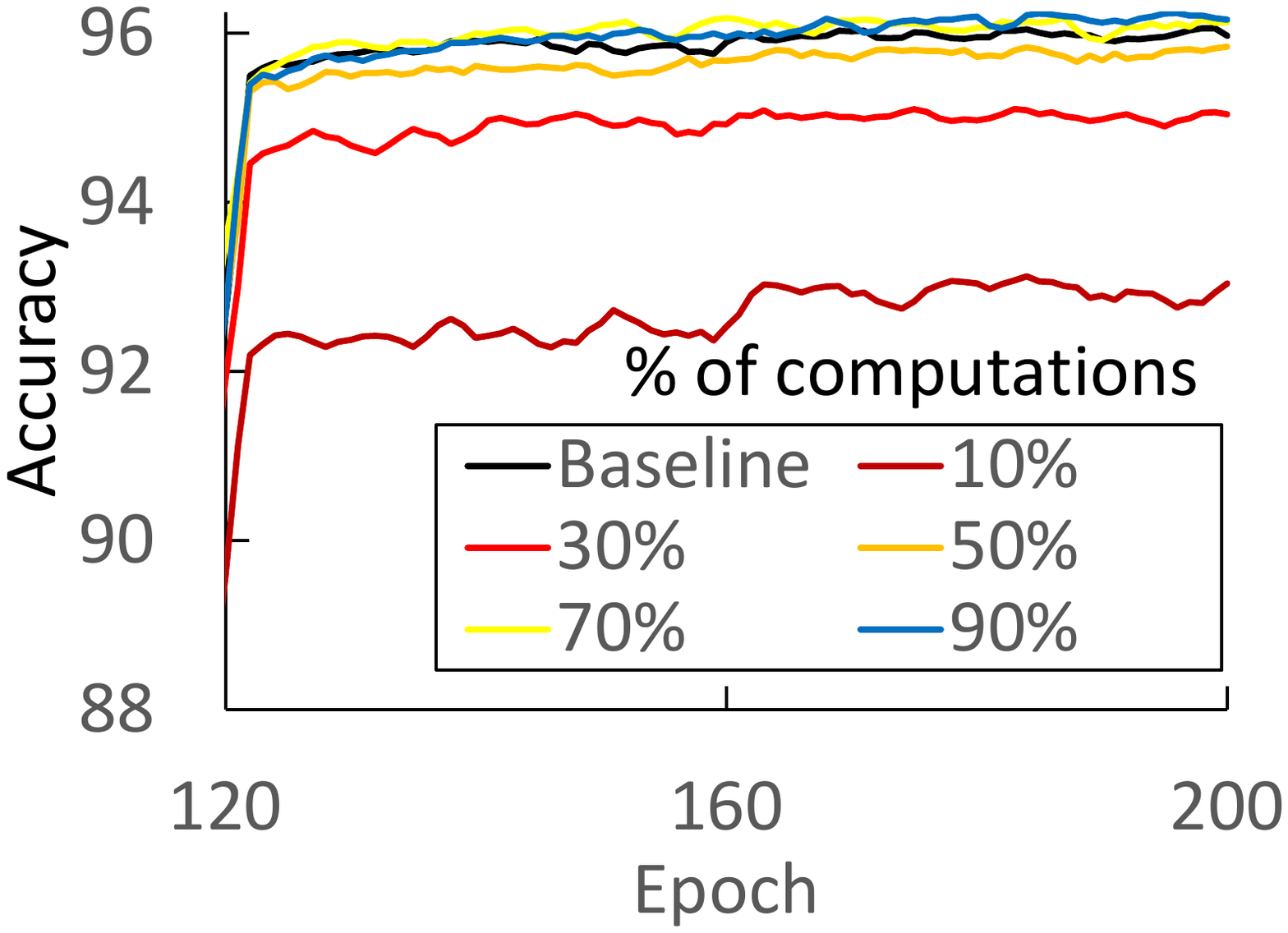}
  \caption{Learning curves for WRN-28-10 CIFAR-10 validation accuracy under different top-$k$ sampling ratios. Focused view of last training epochs}
  \label{fig_sup:wrn_cifar10_learning_last}
\end{figure}

Figure~\ref{fig:wrn_cifar10_learning} shows the CIFAR-10 validation accuracy learning curves for different forward-pass top-$k$ sampling ratios, compared to the non-approximate baseline. We observe that higher sampling ratios lead to slower learning at the early training stages but the gap is decreasing as the training progresses. Figure~\ref{fig_sup:wrn_cifar10_learning_last} focuses on the last training epochs to observe the accuracies in more detail. We observe that 50\% sampling is slightly lower than the non-approximate baseline, while less aggressive approximations that perform 70\% or 90\% of the computations achieve identical or slightly higher validation accuracy.

\subsection{ResNet-50 and ResNet-152 for ImageNet}

The ImageNet \citep{russakovsky2015imagenet} ILSVRC 2012 dataset contain 1.2 million training images of varying dimensions split into 1000 classes. The validation set includes 50K images and the test set consists of 100K images.   

For ResNet-50 \citep{he2016deep} we use the implementation in \url{https://github.com/pytorch/examples/tree/master/imagenet}, available under BSD 3-Clause License. See \citep{he2016deep} for further details on ResNet-50 architecture.

Image preprocessing includes random 224x224 crop, horizontal flipping and image normalization. The optimizer is Momentum SGD with momentum=0.9 and 1e-4 weight decay. Learning rate is 0.1 and it is decayed by 10 every 30 epochs. We use batch size of 256, cross-entropy loss and train the model for 90 epochs. 

We apply sampling to the convolutional layers except the first layer due to the small number of input channels (3) and the fully-connected layer.

\begin{figure}[h]
  \centering
  \subfigure[ResNet-50]{\includegraphics[width=0.5\linewidth]{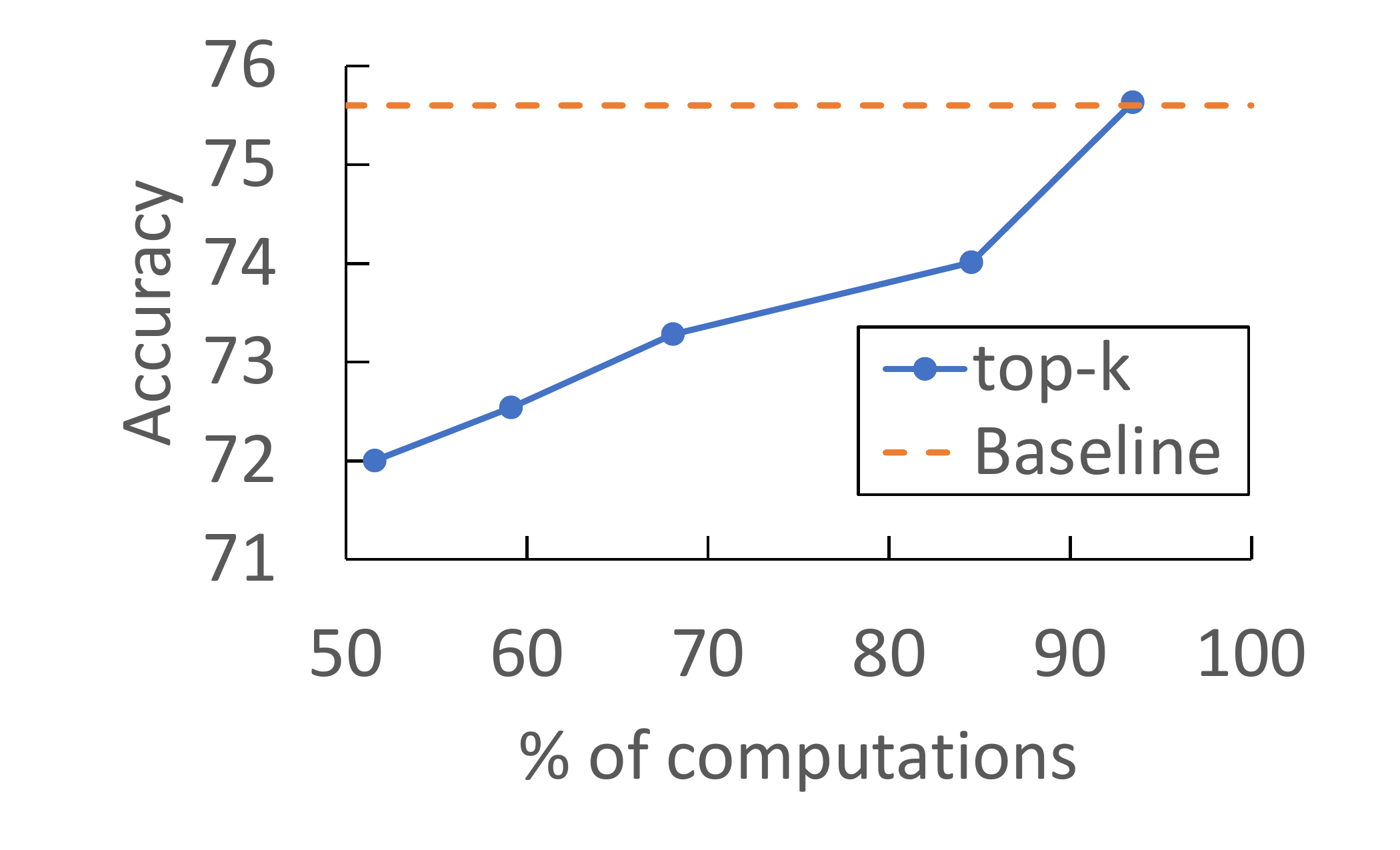}
  \label{fig_sup:resnet50}}~
  \subfigure[ResNet-152]{\includegraphics[width=0.5\linewidth]{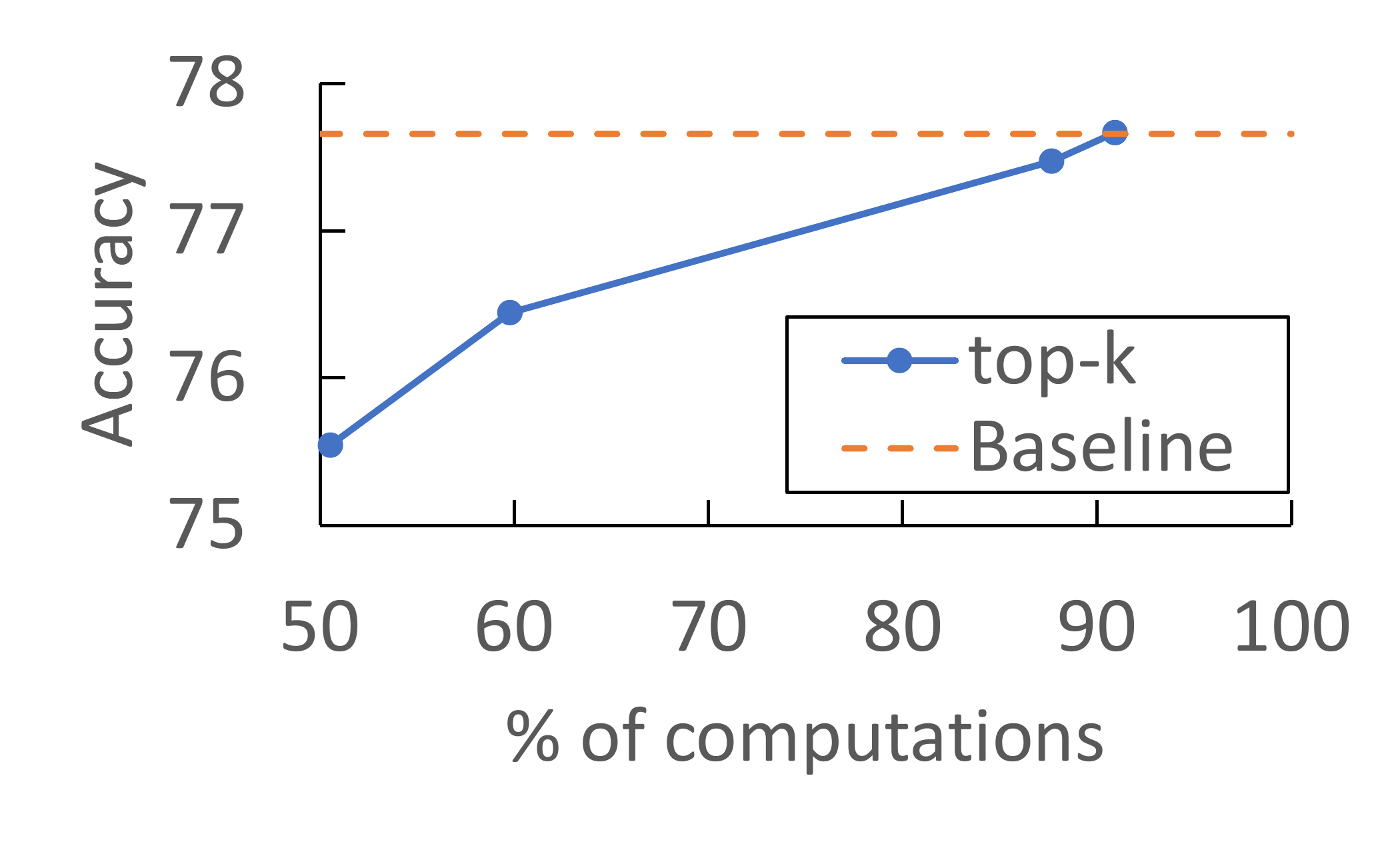}
  \label{fig_sup:resnet152}}
  \caption{ResNet-50 and ResNet-152 ImageNet top-1 test accuracy. The accuracy increases as higher amounts of the computations are performed}
\end{figure}

Figure~\ref{fig_sup:resnet50} shows the top-1 accuracy of ResNet-50 for different sampling ratios. The different data points correspond to 50\% top-$k$ sampling applied to all the layers, all layers with at least 128 channels, 256 channels, 512 channels and 1024 channels.

Figure~\ref{fig_sup:resnet152} shows the top-1 accuracy of ResNet-152 for different sampling ratios. The different data points correspond to 50\% top-$k$ sampling applied to all the layers, all layers with at least 256 channels, 512 channels and 1024 channels.  

Figures~\ref{fig:resnet50_learning} and~\ref{fig:resnet152_learning} show the top-1 validation accuracy learning curves for different forward-pass top-$k$ sampling ratios, compared to the non-approximate baseline. We observe that ResNet-50 and ResNet-152 are more sensitive to sampling compared to WRN-28-10 on CIFAR10. Nonetheless, applying 50\% sampling in the layers with 1024 channels, corresponding to 93\% of the computations in ResNet-50 and 91\% of the computations in ResNet-152, follow the non-approximate learning curves almost identically.

\subsection{Distributed Training}
\begin{figure}[h]
  \centering
  \includegraphics[width=0.5\linewidth]{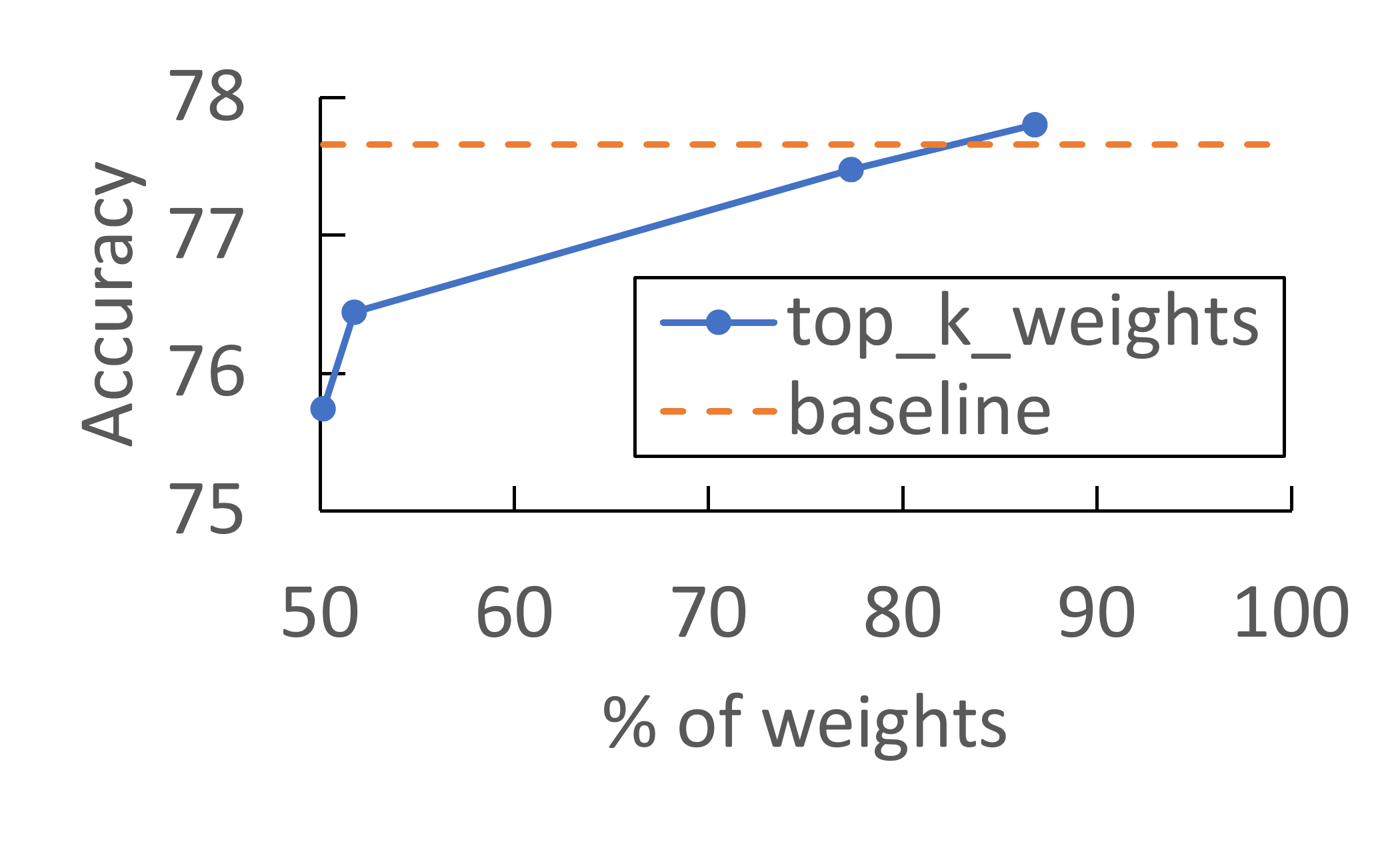}
  \caption{ResNet-152 ImageNet top-1 test accuracy, using top-$k$-weights algorithm.}
  \label{fig_sup:resnet152_topkweights}
\end{figure}

To evaluate the accuracy of top-$k$-weights algorithm for ResNet-152 on Imagenet we used the same settings as in the previous section and trained on a single node with 4 GPUs. The accuracy results are shown in figure \ref{fig_sup:resnet152_topkweights}. The different data points correspond to 50\% top-$k$-weights sampling applied to all layers with at least 256 channels, 512 channels and 1024 channels.  

For the distributed training experiments we used eight AWS EC2 instances equipped with 2.7GHz Intel Xeon E5-2686v4 CPU, one V100 GPU with 16 GB of memory, 10 Gbps networking, PyTorch version 1.7.1, CUDA 11 and Python 3.7.6. 

For the distributed measurement we used the same hyper-parameters except the minibatch size which we set to 32 per GPU. We could not increase the batch size since the AWS EC2 GPU we used had 16GB of memory and could not support higher batch size. We note that the eight-node setting has a total global batch size of 256, which matches the batch size used in the accuracy evaluation.

\end{document}